%% file: arxiv.tex
\documentclass[10pt,twocolumn,letterpaper]{article}

\usepackage[pagenumbers]{cvpr} %
\usepackage{array}
\input{preamble}

\definecolor{cvprblue}{rgb}{0.21,0.49,0.74}
\usepackage[pagebackref,breaklinks,colorlinks,citecolor=cvprblue]{hyperref}

\def\confName{CVPR}
\def\confYear{2024}

\title{\vspace{-40pt} \includegraphics[scale=0.07]{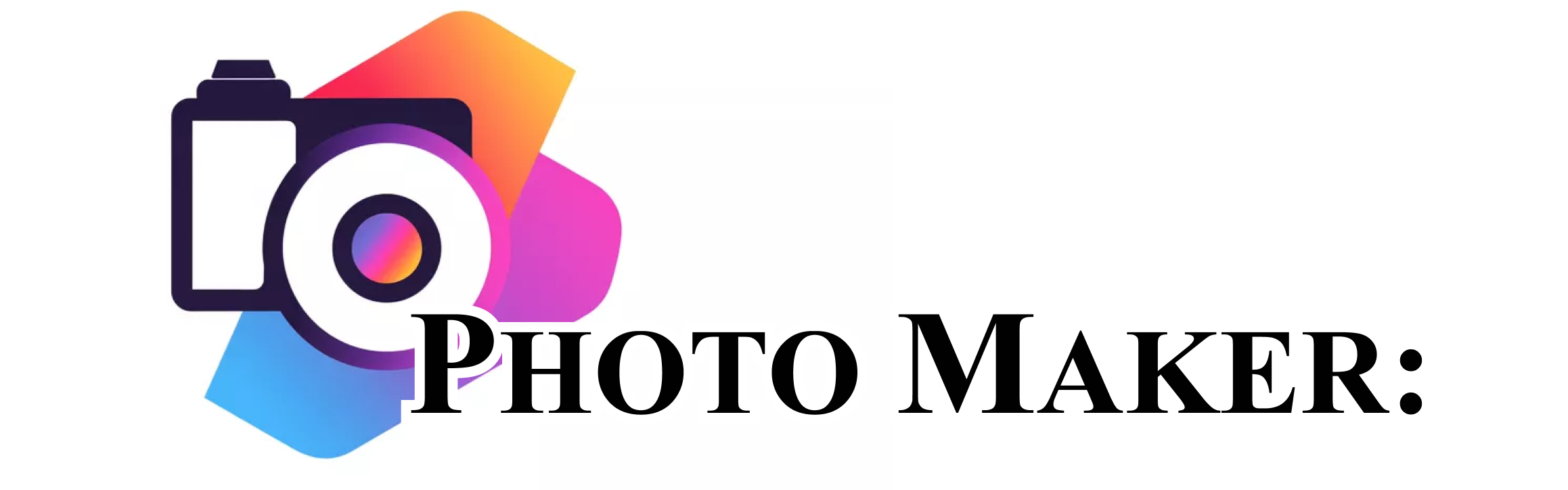} \vspace{-5pt}\\
Customizing Realistic Human Photos via Stacked ID Embedding}

\author{Zhen Li\textsuperscript{1,2 *} \hspace{5pt} Mingdeng Cao\textsuperscript{2,3 *} \hspace{5pt} Xintao Wang\textsuperscript{2 $\dag$} \hspace{5pt} Zhongang Qi\textsuperscript{2} \hspace{5pt} Ming-Ming Cheng\textsuperscript{1 $\dag$} \hspace{5pt} Ying Shan\textsuperscript{2}\\
\textsuperscript{1}VCIP, CS, Nankai University \quad
$^2$ARC Lab, Tencent PCG \quad
$^3$The University of Tokyo\\
\url{https://photo-maker.github.io/}
\vspace{5pt}
}

\begin{document}

\twocolumn[{
\renewcommand\twocolumn[1][]{#1}
\maketitle
\begin{center}
    \centering
    \vspace*{-.8cm}
    \includegraphics[width=\textwidth]{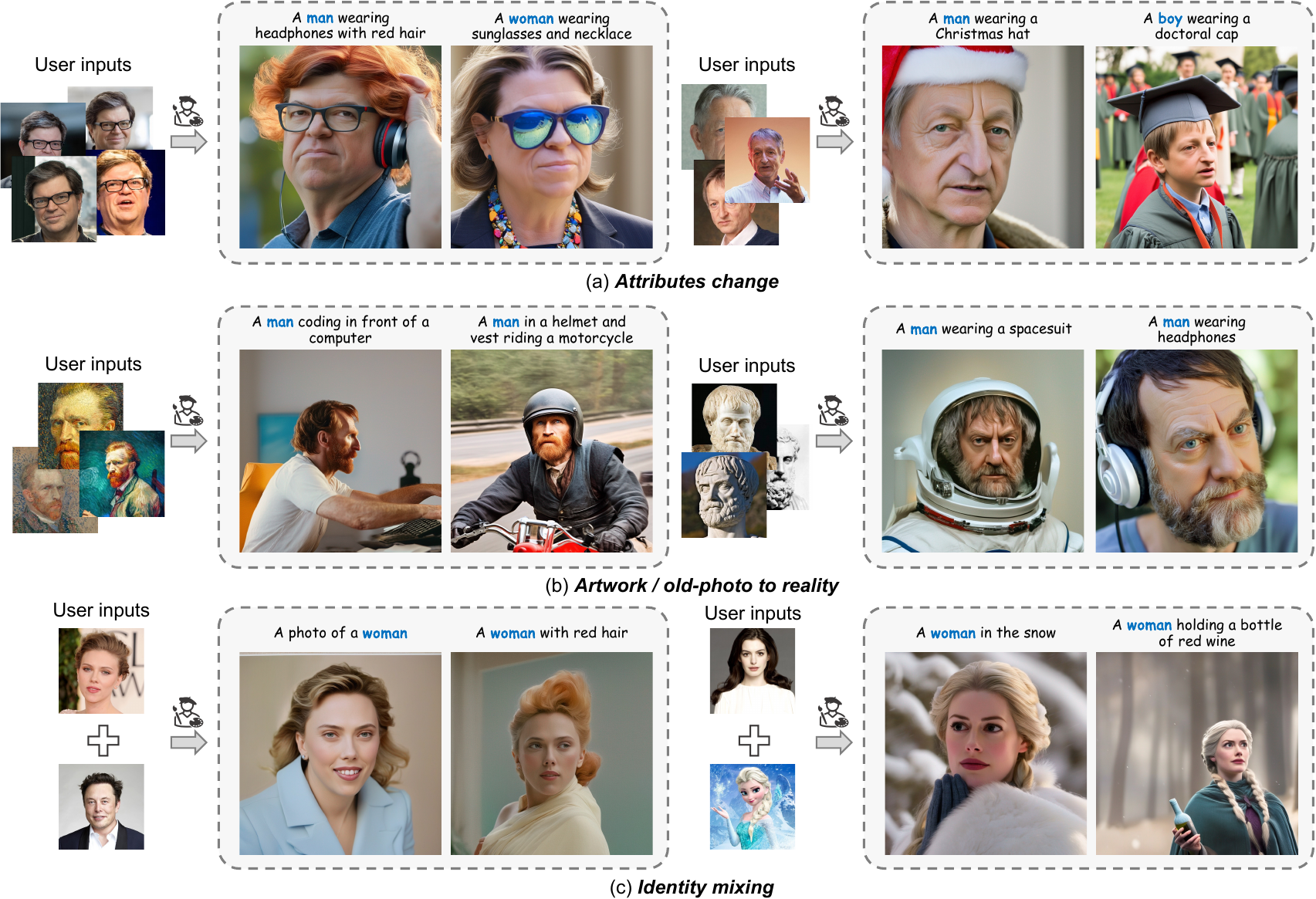}
    \vspace*{-.6cm}
    \captionof{figure}{Given a few images of input ID(s), the proposed \emph{PhotoMaker} can generate diverse personalized ID images based on the text prompt \textbf{in a single forward pass}. Our method can well preserve the ID information from the input image pool while generating realistic human photos. \emph{PhotoMaker} also empowers many interesting applications such as (a) changing attributes, (b) bringing persons from artworks or old photos into reality, or (c) performing identity mixing. (\textit{Zoom-in for the best view})}
\label{fig:teaser}
\end{center}
}]
\let\thefootnote\relax\footnotetext{$^*$ Interns in ARC Lab, Tencent PCG \hspace{3pt} $^\dagger$ Corresponding authors
}

\newcommand{\thefootnote}{\arabic{footnote}}

\input{sec/0_abstract}

\input{sec/1_intro}

\input{sec/2_relatedwork}

\input{sec/3_method}
\input{sec/4_experiments}

\input{sec/5_conclusion}

{
    \small
    \bibliographystyle{ieeenat_fullname}
    \bibliography{main}
}

\input{sec/X_suppl}

\end{document}

%% file: preamble.tex
\usepackage[dvipsnames]{xcolor}

\newcommand{\figref}[1]{Fig.~\ref{#1}}%
\newcommand{\tabref}[1]{Tab.~\ref{#1}}%
\newcommand{\secref}[1]{Sec.~\ref{#1}}
\renewcommand{\eqref}[1]{Eqn.~(\ref{#1})}
\newcommand{\methodname}{PhotoMaker\xspace}
\renewcommand{\paragraph}[1]{\vspace{2pt}\noindent\textbf{#1}}
\newcommand{\sota}[1]{\textbf{#1}}
\newcommand{\subsota}[1]{\underline{#1}}
\newcommand{\tablestyle}[2]{\setlength{\tabcolsep}{#1}\renewcommand{\arraystretch}{#2}\centering\footnotesize}
\definecolor{baselinecolor}{gray}{.9}

\usepackage{circledsteps}
\usepackage{multirow}

%% file: sec/0_abstract.tex
\begin{abstract}
Recent advances in text-to-image generation have made remarkable progress in synthesizing realistic human photos conditioned on given text prompts. 
However, existing personalized generation methods cannot simultaneously satisfy the requirements of high efficiency, promising identity (ID) fidelity, and flexible text controllability.
In this work, we introduce \textbf{PhotoMaker}, an efficient personalized text-to-image generation method, which mainly encodes an arbitrary number of input ID images into a stack ID embedding for preserving ID information.
Such an embedding, serving as a unified ID representation, can not only encapsulate the characteristics of the same input ID comprehensively, but also accommodate the characteristics of different IDs for subsequent integration.
This paves the way for more intriguing and practically valuable applications.
Besides, to drive the training of our PhotoMaker, we propose an ID-oriented data construction pipeline to assemble the training data.
Under the nourishment of the dataset constructed through the proposed pipeline,
our PhotoMaker demonstrates better ID preservation ability than test-time fine-tuning based methods, yet provides significant speed improvements, high-quality generation results, strong generalization capabilities, and a wide range of applications. 
\vspace{-10pt}
\end{abstract}

%% file: sec/1_intro.tex
\section{Introduction}
\label{sec:intro}

Customized image generation related to humans~\cite{ju2023humansd, liu2023hyperhuman, ruiz2023hyperdreambooth} has received considerable attention, giving rise to numerous applications, such as personalized portrait photos~\cite{photoai}, image animation~\cite{zhang2023sadtalker}, and virtual try-on~\cite{wang2018toward}. Early methods~\cite{nitzan2022mystyle, melnik2022face}, limited by the capabilities of generative models (\textit{i.e.}, GANs~\cite{goodfellow2020generative, karras2019style}), could only customize the generation of the facial area, resulting in low diversity, scene richness, and controllability.
Thanks to larger-scale text-image pair training datasets~\cite{schuhmann2022laion}, larger generation models~\cite{saharia2022photorealistic, podell2023sdxl}, and text/visual encoders~\cite{radford2021learning, raffel2020exploring} that can provide stronger semantic embeddings, diffusion-based text-to-image generation models have been continuously evolving recently.
This evolution enables them to generate increasingly realistic facial details and rich scenes.
The controllability has also greatly improved due to the existence of text prompts and structural guidance~\cite{zhang2023adding, mou2023t2i}

Meanwhile, under the nurturing of powerful diffusion text-to-image models, many diffusion-based customized generation algorithms~\cite{ruiz2022dreambooth, gal2022image} have emerged to meet users' demand for high-quality customized results.
The most widely used in both commercial and community applications are DreamBooth-based methods~\cite{ruiz2022dreambooth, db_lora}.
Such applications require dozens of images of the same identity (ID) to fine-tune the model parameters.
Although the results generated have high ID fidelity, there are two obvious drawbacks: one is that customized data used for fine-tuning each time requires manual collection and thus is very time-consuming and laborious; the other is that customizing each ID requires 10-30 minutes, consuming a large amount of computing resources, especially when the generation model becomes larger.
Therefore, to simplify and accelerate the customized generation process, recent works, driven by existing human-centric datasets~\cite{liu2015faceattributes,karras2019style}, have trained visual encoders~\cite{xiao2023fastcomposer, chen2023photoverse} or hypernetworks~\cite{ruiz2023hyperdreambooth, arar2023domain} to represent the input ID images as embeddings or LoRA~\cite{hu2021lora} weights of the model.
After training, users only need to provide an image of the ID to be customized, and personalized generation can be achieved through a few dozen steps of fine-tuning or even without any tuning process.
However, the results customized by these methods cannot simultaneously possess ID fidelity and generation diversity like DreamBooth (see \figref{fig:comp_recontext}). This is because that: 1) during the training process, both the target image and the input ID image sample from the same image. The trained model easily remembers characteristics unrelated to the ID in the image, such as expressions and viewpoints, which leads to poor editability, and 2) relying solely on a single ID image to be customized makes it difficult for the model to discern the characteristics of the ID to be generated from its internal knowledge, resulting in unsatisfactory ID fidelity.

Based on the above two points, and inspired by the success of DreamBooth, in this paper, we aim to:
1) ensure that the ID image condition and the target image exhibit \textit{variations} in viewpoints, facial expressions, and accessories, so that the model does not memorize information that is irrelevant to the ID;
2) provide the model with \textit{multiple different images} of the same ID during the training process to more comprehensively and accurately represent the characteristics of the customized ID.

Therefore, we propose a simple yet effective feed-forward customized human generation framework that can receive multiple input ID images, termed as \textit{\methodname.} 
To better represent the ID information of each input image, we stack the encodings of multiple input ID images at the semantic level, constructing a stacked ID embedding. 
This embedding can be regarded as a unified representation of the ID to be generated, and each of its subparts corresponds to an input ID image.
To better integrate this ID representation and the text embedding into the network, we replace the class word (\textit{e.g.}, man and woman) of the text embedding with the stacked ID embedding. 
The result embedding simultaneously represents the ID to be customized and the contextual information to be generated.
Through this design, without adding extra modules in the network, the cross-attention layer of the generation model itself can adaptively integrate the ID information contained in the stacked ID embedding.

At the same time, the stacked ID embedding allows us to accept any number of ID images as input during inference while maintaining the efficiency of the generation like other tuning-free methods~\cite{shi2023instantbooth, xiao2023fastcomposer}. 
Specifically, our method requires about 10 seconds to generate a customized human photo when receiving four ID images, which is about 130$\times$  faster than DreamBooth\footnote{Test on one NVIDIA Tesla V100}.
Moreover, since our stacked ID embedding can represent the customized ID more comprehensively and accurately, our method can provide better ID fidelity and generation diversity compared to state-of-the-art tuning-free methods. 
Compared to previous methods, our framework has also greatly improved in terms of controllability. 
It can not only perform common recontextualization but also change the attributes of the input human image (\textit{e.g.}, accessories and expressions), generate a human photo with completely different viewpoints from the input ID, and even modify the input ID's gender and age (see \figref{fig:teaser}).

It is worth noticing that our PhotoMaker 
also unleashes a lot of possibilities for users to generate customized human photos. 
Specifically, although the images that build the stacked ID embedding come from the same ID during training, we can use different ID images to form the stacked ID embedding during inference to merge and create a new customized ID. 
The merged new ID can retain the characteristics of different input IDs. 
For example, we can generate Scarlett Johansson that looks like  Elun Musk or a customized ID that mixes a person with a well-known IP character (see \figref{fig:teaser}(c)).
At the same time, the merging ratio can be simply adjusted by prompt weighting~\cite{hertz2022prompt, prompt_weighting} or by changing the proportion of different ID images in the input image pool, demonstrating the flexibility of our framework.

Our PhotoMaker necessitates the simultaneous input of multiple images with the same ID during the training process, thereby requiring the support of an ID-oriented human dataset.
However, existing datasets either do not classify by IDs~\cite{karras2019style, schuhmann2022laion, zheng2022general, liu2023hyperhuman} or only focus on faces without including other contextual information~\cite{liu2015faceattributes, kaisiyuan2020mead, nitzan2022mystyle}.
We therefore design an automated pipeline to construct an ID-related dataset to facilitate the training of our \methodname. 
Through this pipeline, we can build a dataset that includes a large number of IDs, each with multiple images featuring diverse viewpoints, attributes, and scenarios. 
Meanwhile, in this pipeline, we can automatically generate a caption for each image, marking out the corresponding class word~\cite{ruiz2022dreambooth}, to better adapt to the training needs of our framework.

%% file: sec/2_relatedwork.tex
\section{Related work}
\paragraph{Text-to-Image Diffusion Models.}
Diffusion models~\cite{ho2020denoising, song2020denoising} have made remarkable progress in text-conditioned image generation~\cite{ saharia2022photorealistic, rombach2022high, kawar2022imagic, ramesh2022hierarchical}, attracting widespread attention in recent years.
The remarkable performance of these models can be attributable to  high-quality large-scale text-image datasets~\cite{changpinyo2021conceptual,schuhmann2021laion,schuhmann2022laion}, the continuous upgrades of foundational models~\cite{chen2023pixart, peebles2023scalable}, conditioning 
encoders~\cite{radford2021learning,ilharco_gabriel_2021_5143773, raffel2020exploring}, and the improvement of controllability~\cite{li2023gligen, zhang2023adding, mou2023t2i, ye2023ip}. 
Due to these advancements,  Podell~\etal~\cite{podell2023sdxl} developed the currently most powerful open-source generative model, SDXL.
Given its impressive capabilities in generating human portraits, we build our PhotoMaker based on this model. 
However, our method can also be extended to other text-to-image synthesis models.

\paragraph{Personalization in Diffusion Models.}
Owing to the powerful generative capabilities  of the diffusion models, more researchers try to explore personalized generation based on them.
Currently, mainstream personalized synthesis methods can be mainly divided into two categories. One relies on additional optimization during the test phase, such as DreamBooth~\cite{ruiz2022dreambooth} and Textual Inversion~\cite{gal2022image}. 
Given that both pioneer works require substantial time for fine-tuning, some studies have attempted to expedite the process of personalized customization by reducing the number of parameters needed for tuning~\cite{kumari2022multi, db_lora, han2023svdiff, yuan2023inserting} or by pre-training with large datasets~\cite{gal2023designing, ruiz2023hyperdreambooth}.
 Despite these advances, they still require extensive fine-tuning of the pre-trained model for each new concept, making the process time-consuming and restricting its applications.
 Recently, some studies~\cite{wei2023elite, ma2023unified, jia2023taming, chen2023subject, shi2023instantbooth, ma2023subject, chen2023anydoor} attempt to perform personalized generation using a single image with a single forward pass, significantly accelerating the personalization process.
These methods either utilize personalization datasets~\cite{chen2023subject, sohn2023styledrop} for training~ or encode the images to be customized  in the semantic space~\cite{wei2023elite, ma2023unified, jia2023taming, 
chen2023photoverse,
shi2023instantbooth, xiao2023fastcomposer}.
Our method focuses on the generation of human portraits based on both of the aforementioned technical approaches.
Specifically,
it not only relies on the construction of an ID-oriented personalization dataset, but also on obtaining the embedding that represents the person's ID in the semantic space.
Unlike previous embedding-based methods, our PhotoMaker extracts a stacked ID embedding from multiple ID images. 
While providing better ID representation, the proposed method can maintain the same high efficiency as previous embedding-based methods.

%% file: sec/3_method.tex
\section{Method}
\begin{figure*}[t]
  \centering
  \includegraphics[width=0.95\textwidth]{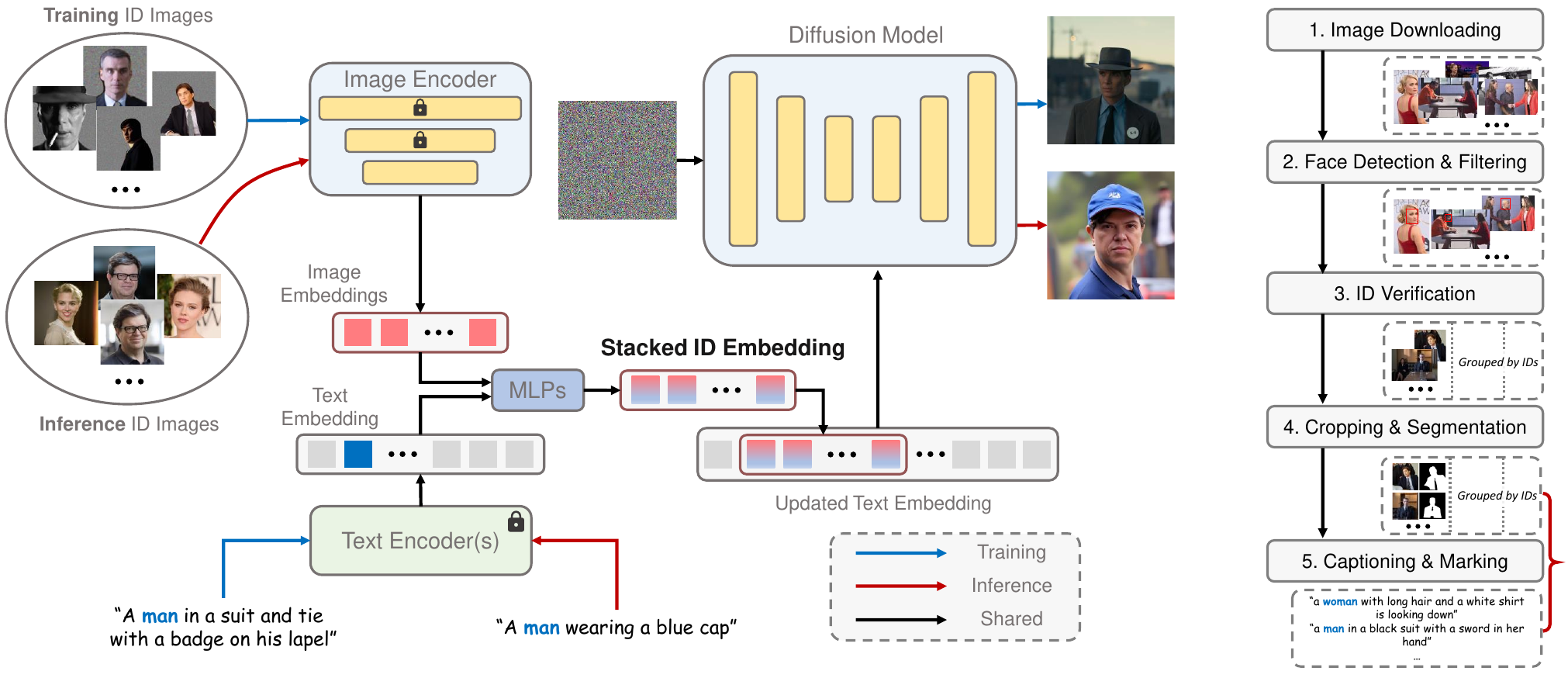}
  \vspace{-10pt}
  \tablestyle{5pt}{1.15}
    \begin{center}
    \begin{tabular}{>{\centering\arraybackslash}p{0.75\textwidth}>{\centering\arraybackslash}p{0.2\textwidth}}
    (a) & (b) \\
    \end{tabular}
    \end{center}
  \vspace{-15pt}
  \caption{\textbf{Overviews of the proposed (a) \textit{PhotoMaker} and (b) ID-oriented data construction pipeline}.
  For the proposed PhotoMaker, we first obtain the text embedding and image embeddings from text encoder(s) and image encoder, respectively.
  Then, we extract the fused embedding by merging the corresponding class embedding (\eg, \textit{man} and \textit{woman}) and each image embedding.
  Next, we concatenate all fused embeddings along the length dimension to form the \textit{stacked ID embedding}.
  Finally, we feed the stacked ID embedding to all cross-attention layers for adaptively merging the ID content in the diffusion model.
Note that although we use images of the same ID with the masked background during training, we can directly input images of different IDs without background distortion to create a new ID during inference.}
  \label{fig:overview}
  \vspace{-10pt}
\end{figure*}

\subsection{Overview}

Given a few ID images to be customized, the goal of our PhotoMaker is to generate a new photo-realistic human image that retains the characteristics of the input IDs and changes the content or the attributes of the generated ID under the control of the text prompt. 
Although we input multiple ID images for customization like DreamBooth, we still enjoy the same efficiency as other tuning-free methods, accomplishing customization with a single forward pass, while maintaining promising ID fidelity and text edibility. 
In addition, we can also mix multiple input IDs, and the generated image can well retain the characteristics of different IDs, which releases possibilities for more applications. 
The above capabilities are mainly brought by our proposed simple yet effective \textit{stacked ID embedding}, which can provide a unified representation of the input IDs. 
Furthermore, to facilitate training our PhotoMaker, we design a data construction pipeline to build a human-centric dataset classified by IDs.
\figref{fig:overview}(a) shows the overview of the proposed PhotoMaker.
\figref{fig:overview}(b) shows our data construction pipeline.

\subsection{Stacked ID Embedding}
\label{sec:embedding}
\paragraph{Encoders.} 
Following recent works~\cite{jia2023taming, wei2023elite, shi2023instantbooth}, we use the CLIP~\cite{radford2021learning} image encoder $\mathcal{E}_{img}$ to extract image embeddings for its alignment with the original text representation space in diffusion models.
Before feeding each input image into the image encoder, we filled the image areas other than the body part of a specific ID with random noises to eliminate the influence of other IDs and the background.
Since the data used to train the original CLIP image encoder mostly consists of natural images, to better enable the model to extract ID-related embeddings from the masked images, we finetune part of the transformer layers in the image encoder when training our PhotoMaker.
We also introduce additional learnable projection layers to inject the embedding obtained from the image encoder into the same dimension as the text embedding.
Let $\{X^{i}  \mid i=1 \dots N\}$ denote $N$ input ID images acquired from a user,
we thus obtain the extracted embeddings $\{e^{i} \in \mathbb{R}^{D} \mid i=1 \dots N\}$, where $D$ denotes the projected dimension.
Each embedding corresponds to the ID information of an input image.
For a given text prompt $T$, we extract text embeddings $ t \in \mathbb{R}^{L \times D} $ using the pre-trained CLIP text encoder $\mathcal{E}_{text}$, where $L$ denotes the length of the embedding.

\paragraph{Stacking.} 
Recent works~\cite{ruiz2022dreambooth, gal2022image, xiao2023fastcomposer} have shown that, in the text-to-image models, personalized character ID information can be represented by some \textit{unique tokens}. 
Our method also has a similar design to better represent the ID information of the input human images. 
Specifically, we mark the corresponding class word (\eg, \textit{man} and \textit{woman}) in the input caption (see \secref{sec:data_pipeline}).
We then extract the feature vector at the corresponding position of the class word in the text embedding.
This feature vector will be fused with each image embedding $e^{i}$.
We use two MLP layers to perform such a fusion operation.
The fused embeddings can be denoted as $\{\hat{e}^{i} \in \mathbb{R}^{D} \mid i=1 \dots N\}$.
By combining the feature vector of the class word, this embedding can represent the current input ID image more comprehensively.
In addition, during the inference stage, this fusion operation also provides stronger semantic controllability for the customized generation process. For example, we can customize the age and gender of the human ID by simply replacing the class word (see \secref{sec:applications}).

After obtaining the fused embeddings, we concatenate them along the length dimension to form the \textit{stacked id embedding}:
\begin{equation}
    s^{*} = \mathtt{Concat}([\hat{e}^{1}, \dots, \hat{e}^{N}]) \quad s^{*} \in \mathbb{R}^{N\times D}.
\label{eq:stacked_embedding}
\end{equation}
This stacked ID embedding can serve as a unified representation of multiple ID images while it retains the original representation of each input ID image.
It can accept any number of ID image encoded embeddings, therefore, its length $N$ is variable. 
Compared to DreamBooth-based methods~\cite{ruiz2022dreambooth,db_lora}, which inputs multiple \textit{images} to finetune the model for personalized customization, our method essentially sends multiple \textit{embeddings} to the model simultaneously. 
After packaging the multiple images of the same ID into a batch as the input of the image encoder, a stacked ID embedding can be obtained through a single forward pass, significantly enhancing efficiency compared to tuning-based methods.
Meanwhile, compared to other embedding-based methods~\cite{wei2023elite, xiao2023fastcomposer}, this unified representation can maintain both promising ID fidelity and text controllability, as it contains more comprehensive ID information.
In addition, it is worth noting that, although we only used multiple images of the same ID to form this stacked ID embedding during training, we can use images that come from different IDs to construct it during the inference stage.
Such flexibility opens up possibilities for many interesting applications.
For example, we can mix two persons that exist in reality or mix a person and a well-known character IP (see \secref{sec:applications}).

\paragraph{Merging.}
We use the inherent cross-attention mechanism in diffusion models to adaptively merge the ID information contained in stacked ID embedding.
We first replace the feature vector at the position corresponding to the class word in the original text embedding $t$ with the stacked id embedding $s^{*}$, resulting in an update text embedding $t^{*} \in \mathbb{R}^{(L+N-1)\times D}$.
Then, the cross-attention operation can be formulated as:

\begin{equation}
\left \{
\begin{array}{ll}
    \mathbf{Q} = \mathbf{W}_Q\cdot \phi(z_t);\ \mathbf{K} = \mathbf{W}_K\cdot t^{*};\ \mathbf{V} = \mathbf{W}_V\cdot t^{*}\\
    \mathtt{Attention}(\mathbf{Q}, \mathbf{K}, \mathbf{V}) = \mathtt{softmax}(\frac{\mathbf{Q}\mathbf{K}^T}{\sqrt{d}})\cdot \mathbf{V},
\end{array}
\right.
\label{eq:cross-attn}
\end{equation}
where $\phi(\cdot)$ is an embedding that can be encoded from the input latent by the UNet denoiser. $\mathbf{W}_Q$, $\mathbf{W}_K$, and $\mathbf{W}_V$ are projection matrices.
Besides, we can adjust the degree of participation of one input ID image in generating the new customized ID through prompt weighting~\cite{hertz2022prompt, prompt_weighting}, demonstrating the \textit{flexibility} of our PhotoMaker.
Recent works~\cite{kumari2022multi, db_lora} found that good ID customization performance can be achieved by simply tuning the weights of the attention layers.
To make the original diffusion models better perceive the ID information contained in stacked ID embedding, we additionally train the LoRA~\cite{hu2021lora, db_lora}  residuals of the matrices in the attention layers.

\subsection{ID-Oriented Human Data Construction}
\label{sec:data_pipeline}

Since our PhotoMaker needs to sample multiple images of the same ID for constructing the stacked ID embedding during the training process, we need to use a dataset classified by IDs to drive the training process of our PhotoMaker.
However, existing human datasets either do not annotate ID information~\cite{karras2019style, schuhmann2022laion, zheng2022general, liu2023hyperhuman}, or the richness of the scenes they contain is very limited~\cite{liu2015faceattributes, kaisiyuan2020mead, nitzan2022mystyle} (\ie, they only focus on the face area).
Thus, in this section, we will introduce a pipeline for constructing a human-centric text-image dataset, which is classified by different IDs. 
\figref{fig:overview}(b) illustrates the proposed pipeline.
Through this pipeline, we can collect an ID-oriented dataset, which contains a large number of IDs, and each ID has multiple images that include different expressions, attributes, scenes, etc. 
This dataset not only facilitates the training process of our PhotoMaker but also may inspire potential future ID-driven research.
The statistics of the dataset are shown in the appendix.

\paragraph{Image downloading.}  
We first list a roster of celebrities, which can be obtained from VoxCeleb1 and VGGFace2~\cite{cao2018vggface2}. 
We search for names in the search engine according to the list and crawled the data. 
About 100 images were downloaded for each name. 
To generate higher quality portrait images~\cite{podell2023sdxl}, we filtered out images with the shortest side of the resolution less than 512 during the download process.

\paragraph{Face detection and filtering.} 
We first use RetinaNet~\cite{deng2020retinaface} to detect face bounding boxes and filter out the detections with small sizes (less than 256 × 256). 
If an image does not contain any bounding boxes that meet the requirements, the image will be filtered out.
We then perform ID verification for the remaining images.

\paragraph{ID verification.} 
Since an image may contain multiple faces, we need first to identify which face belongs to the current identity group.
Specifically, we send all the face regions in the detection boxes of the current identity group into ArcFace~\cite{deng2019arcface} to extract identity embeddings and calculate the L2 similarity of each pair of faces.
We sum the similarity calculated by each identity embedding with all other embeddings to get the score for each bounding box. 
We select the bounding box with the highest sum score for each image with multiple faces.
After bounding box selection, we recompute the sum score for each remaining box. 
We calculate the standard deviation $\delta$ of the sum score by ID group. 
We empirically use $8\delta$ as a threshold to filter out images with inconsistent IDs.

\paragraph{Cropping and segmentation.}
We first crop the image with a larger square box based on the detected face area while ensuring that the facial region can occupy more than 10\% of the image after cropping.
Since we need to remove the irrelevant background and IDs from the input ID image before sending it into the image encoder, we need to generate the mask for the specified ID.
Specifically, we employ the Mask2Former~\cite{cheng2021mask2former} to perform panoptic segmentation for the `person' class.
We leave the mask with the highest overlap with the facial bounding box corresponding to the ID.
Besides, we choose to discard images where the mask is not detected, as well as images where no overlap is found between the bounding box and the mask area.

\paragraph{Captioning and marking}
We use BLIP2~\cite{li2023blip} to generate a caption for each cropped image.
Since we need to mark the class word (\eg,  man, woman, and boy) to facilitate the fusion of text and image embeddings, we regenerate captions that do not contain any class word using the random mode of BLIP2 until a class word appears.
After obtaining the caption, we singularize the class word in the caption to focus on a single ID. 
Next, we need to mark the position of the class word that corresponds to the current ID. 
Captions that contain only one class word can be directly annotated. 
For captions that contain multiple class words, we count the class words contained in the captions for each identity group. 
The class word with the most occurrences will be the class word for the current identity group.
We then use the class word of each identity group to match and mark each caption in that identity group.
For a caption that does not include the class word that matches that of the corresponding identity group, we employ a dependence parsing model~\cite{montani20212} to segment the caption according to different class words. 
We calculate the CLIP score~\cite{radford2021learning} between the sub-caption after segmentation and the specific ID region in the image.
Besides, we calculate the label similarity between the class word of the current segment and the class word of the current identity group through SentenceFormer~\cite{reimers2019sentence}. 
We choose to mark the class word corresponding to the maximum product of the CLIP score and the label similarity.

%% file: sec/4_experiments.tex
\section{Experiments}

\subsection{Setup}

\paragraph{Implementation details.}
To generate more photo-realistic human portraits, we employ SDXL model~\cite{podell2023sdxl}  \texttt{stable-diffusion-xl-base-1.0} as our text-to-image synthesis model.
Correspondingly, the resolution of training data is resized to $1024 \times 1024$.
We employ CLIP ViT-L/14~\cite{radford2021learning} and an additional projection layer to obtain the initial image embeddings $e^{i}$.
For text embeddings, we keep the original two text encoders in SDXL for extraction.
The overall framework is optimized with Adam~\cite{kingma2014adam}  on 8 NVIDIA A100 GPUs for two weeks 
with a batch size of 48. 
We set the learning rate as $1e-4$ for LoRA weights, and $1e-5$ for other trainable modules.
During training, we randomly sample 1-4  images with the same ID as the current target ID image to form a stacked ID embedding.
Besides, to improve the generation performance by using classifier-free guidance, we have a 10\% chance of using null-text embedding to replace the original updated text embedding $t^{*}$.
We also use masked diffusion loss~\cite{avrahami2023break} with a probability of 50\% to encourage the model to generate more faithful ID-related areas.
During the inference stage, we use delayed subject conditioning~\cite{xiao2023fastcomposer} to solve the conflicts between text and ID conditions.
We use 50 steps of DDIM sampler~\cite{song2020denoising}.
The scale of classifier-free guidance is set to 5.

\paragraph{Evaluation metrics.}
Following DreamBooth~\cite{ruiz2022dreambooth},
we use DINO~\cite{caron2021emerging} and CLIP-I~\cite{gal2022image} metrics to measure the ID fidelity and use CLIP-T~\cite{radford2021learning} metric to measure the prompt fidelity.
For a more comprehensive evaluation,
we also compute the face similarity by detecting and cropping the facial regions between the generated image and the real image with the same ID.
We use RetinaFace~\cite{deng2020retinaface} as the detection model.
Face embedding is extracted by FaceNet~\cite{schroff2015facenet}.
To evaluate the quality of the generation, we employ the FID metric~\cite{heusel2017gans, parmar2021cleanfid}.
Importantly, as most embedding-based methods tend to incorporate facial pose and expression into the representation, the generated images often lack variation in the facial region.
Thus, we propose a metric, named \textit{Face Diversity}, to measure the diversity of the generated facial regions.
Specifically, we first detect and crop the face region in each generated image.
Next, we calculate the LPIPS~\cite{zhang2018perceptual} scores between each pair of facial areas for all generated images and take the average. 
The larger this value, the higher the diversity of the generated facial area.

\paragraph{Evaluation dataset.}
Our evaluation dataset includes 25 IDs, which consist of 9 IDs from Mystyle~\cite{nitzan2022mystyle} and an additional 16 IDs that we collected by ourselves.
Note that these IDs do not appear in the training set, serving to evaluate the generalization ability of the model.
To conduct a more comprehensive evaluation, we also prepare 40 prompts, which cover a variety of expressions, attributes, decorations, actions, and backgrounds.
For each prompt of each ID, we generate 4 images for evaluation.
More details are listed in the appendix.

\begin{figure*}[t]
  \centering
\includegraphics[width=0.95\textwidth]{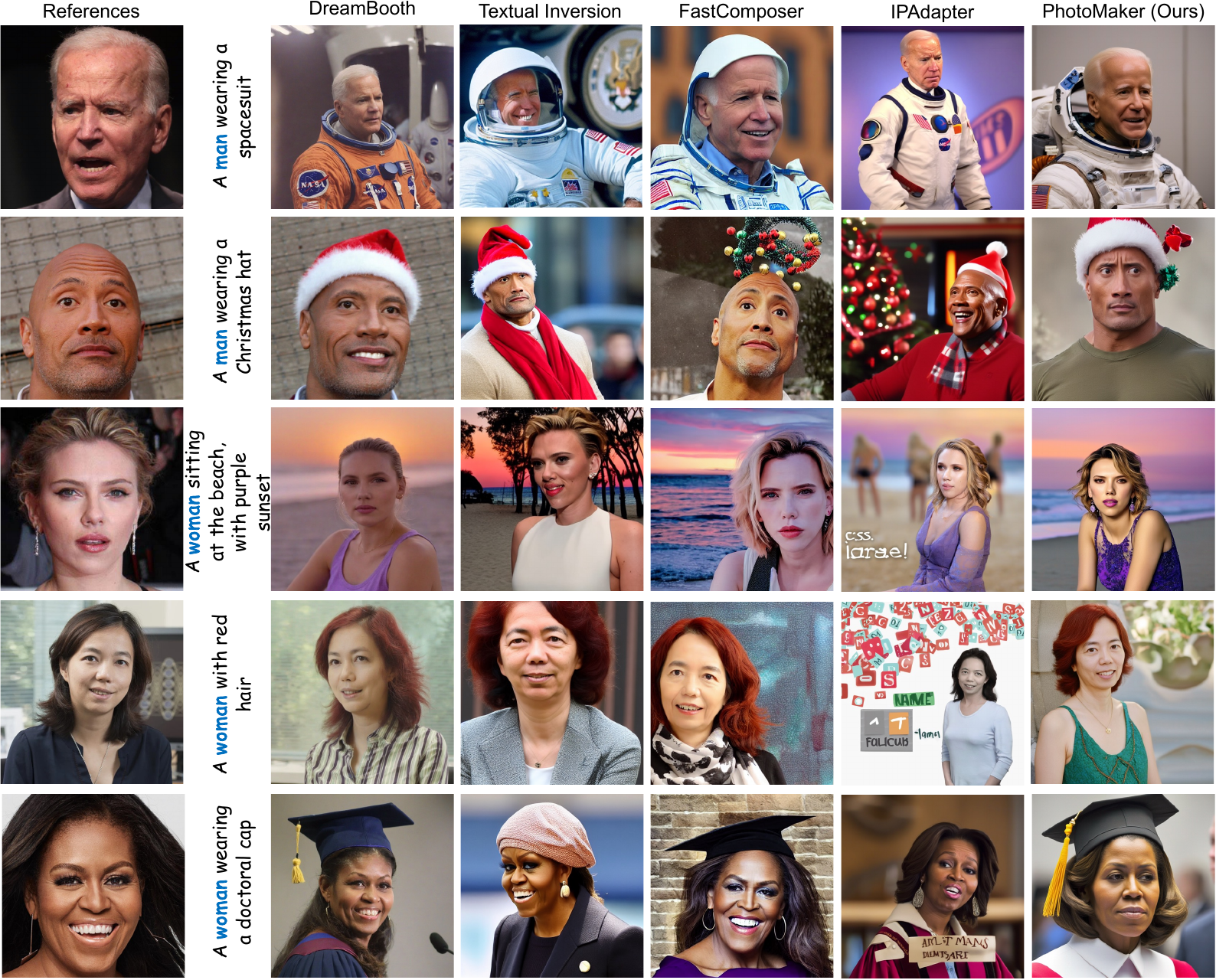}
\vspace{-5pt}
  \caption{\textbf{Qualitative comparison on universal recontextualization samples}. We compare our method with DreamBooth~\cite{ruiz2022dreambooth}, Textual Inversion~\cite{gal2022image}, FastComposer~\cite{xiao2023fastcomposer}, and IPAdapter~\cite{ye2023ip} for five different identities and corresponding prompts. 
  We observe that our method generally achieves high-quality generation, promising editability, and strong identity fidelity. (\textit{Zoom-in for the best view})}
  \label{fig:comp_recontext}
\vspace{-5pt}
\end{figure*}

\begin{table*}
    \tablestyle{5pt}{1.15}
    \centering
    \begin{tabular}{lccccccc} 
        \toprule
         &  CLIP-T$\uparrow$ (\%)&  CLIP-I$\uparrow$ (\%)&  DINO$\uparrow$ (\%)&  Face Sim.$\uparrow$ (\%)&  Face Div.$\uparrow$ (\%)&  FID$\downarrow$&  Speed$\downarrow$ (s)\\ 
        \midrule
         DreamBooth~\cite{ruiz2022dreambooth}&  \sota{29.8}&  62.8&  
39.8&  49.8&  49.1&   374.5& 1284\\ 
         Textual Inversion~\cite{gal2022image}&  24.0&  70.9&  39.3&  54.3&  \sota{59.3}&   \sota{363.5}&  2400\\ 
         FastComposer~\cite{xiao2023fastcomposer}&  \subsota{28.7}&  66.8&  40.2&  61.0&   45.4&   375.1& 8\\ 
         IPAdapter~\cite{ye2023ip}&  25.1&  \subsota{71.2}&  \subsota{46.2}&  \sota{67.1}&  52.4&   375.2& 12\\ 
         PhotoMaker (Ours)&  26.1&  \sota{73.6}&  \sota{51.5}&  \subsota{61.8}&  \subsota{57.7}&   \subsota{370.3}& 10\\ 
        \bottomrule
    \end{tabular}
    \caption{
    \textbf{Quantitative comparison on the universal recontextualization setting}.
    The metrics used for benchmarking cover the ability to preserve ID information (\ie, CLIP-I, DINO, and Face Similarity), text consistency (\ie, CLIP-T), diversity of generated faces (\ie, Face Diversity), and generation quality (\ie, FID).
    Besides, we define personalized speed as the time it takes to obtain the final personalized image after feeding the ID condition(s).
We measure personalized time on a single NVIDIA Tesla V100 GPU.
The best result is shown in \sota{bold}, and the second best is \subsota{underlined}.
}
    \label{tab:comp_recontext}
\vspace{-10pt}
\end{table*}

\begin{figure*}[t]
  \centering  \includegraphics[width=0.95\textwidth]{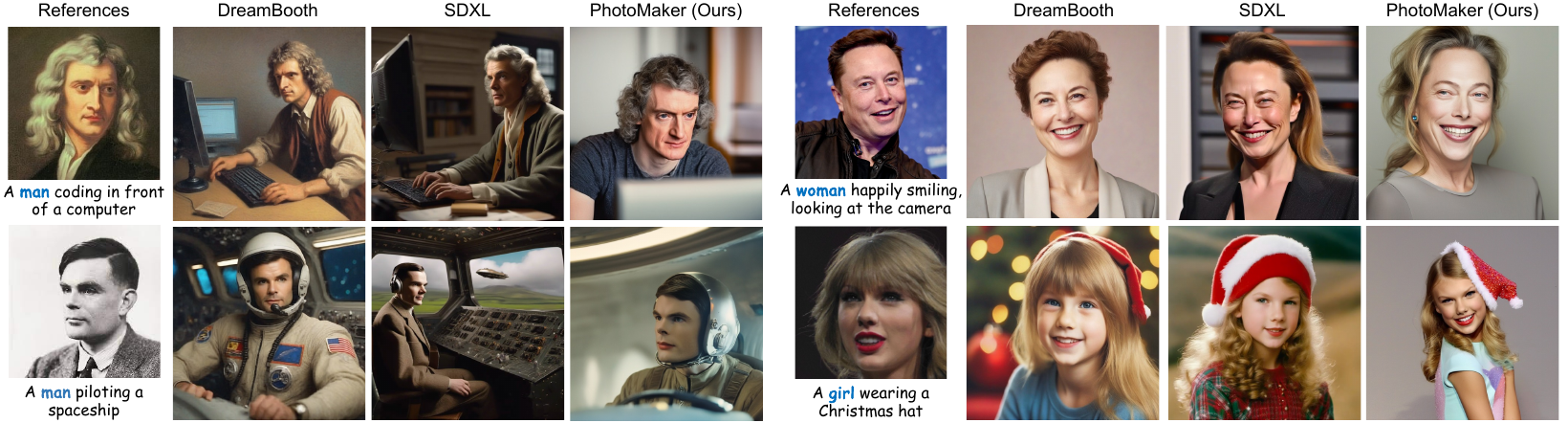}
  \vspace{-10pt}
  \tablestyle{5pt}{1.15}
    \begin{center}
    \begin{tabular}{>{\centering\arraybackslash}p{0.45\textwidth}>{\centering\arraybackslash}p{0.45\textwidth}}
    (a) & \quad (b) \\
    \end{tabular}
    \end{center}
  \vspace{-17pt}
  \caption{\textbf{Applications on (a) artwork and old photo, and (b) changing age or gender.}
  We are able to bring the past people back to real life or change the age and gender of the input ID.
  For the first application,
  we prepare a prompt template \texttt{A photo of <original prompt>, photo-realistic} for DreamBooth and SDXL. 
  Correspondingly, we change the class word to the celebrity name in the original prompt. 
  For the second one, 
   we replace the class word to  \texttt{<class word> <name>, (at the age of 12)} for them.}
  \label{fig:comp_old_age}
  \vspace{-10pt}
\end{figure*}

\subsection{Applications}
\label{sec:applications}

In this section, we will elaborate on the applications that our PhotoMaker can empower.
For each application, we choose the comparison methods which may be most suitable for the corresponding setting.
The comparison method will be chosen from DreamBooth~\cite{ruiz2022dreambooth}, Textual Inversion~\cite{gal2022image}, FastComposer~\cite{xiao2023fastcomposer}, and IPAdapter~\cite{ye2023ip}.
We prioritize using the official model provided by each method.
For DreamBooth and IPAdapter, we use their SDXL versions for a fair comparison.
For all applications, we have chosen four input ID images to form the stacked ID embedding in our PhotoMaker.
We also fairly use four images to train the methods that need test-time optimization.
We provide more samples in the appendix for each application.

\paragraph{Recontextualization}
We first show results with simple context changes such as modified hair color and clothing, or generate backgrounds based on basic prompt control. 
Since all methods can adapt to this application, we conduct quantitative and qualitative comparisons of the generated results (see~\tabref{tab:comp_recontext} and  \figref{fig:comp_recontext}). The results show that our method can well satisfy the ability to generate high-quality images, while ensuring high ID fidelity (with the largest CLIP-T and DINO scores, and the second best Face Similarity). 
Compared to most methods, our method generates images of higher quality, and the generated facial regions exhibit greater diversity.
At the same time, 
our method can maintain a high efficiency consistent with embedding-based methods.
For a more comprehensive comparison, we show the user study results in \secref{sec:user_study} in the appendix.

\paragraph{Bringing person in artwork/old photo into reality.} 
By taking artistic paintings, sculptures, or old photos of a person as input, our PhotoMaker can bring a person from the last century or even ancient times to the present century to ``take" photos for them.
\figref{fig:comp_old_age}(a) illustrate the results.
Compared to our method, both Dreambooth and SDXL have difficulty generating realistic human images that have not appeared in real photos.
In addition, due to the excessive reliance of DreamBooth on the quality and resolution of customized images, it is difficult for DreamBooth to generate high-quality results when using old photos for customized generation.

\paragraph{Changing age or gender.} 
By simply replacing class words (\eg man and woman), our method can achieve changes in gender and age.
\figref{fig:comp_old_age}(b) shows the results.
Although SDXL and DreamBooth can also achieve the corresponding effects after prompt engineering, our method can more easily capture the characteristic information of the characters due to the role of the stacked ID embedding. 
Therefore, our results show a higher ID fidelity.

\begin{figure}[t]
  \centering
  \includegraphics[width=0.45\textwidth]{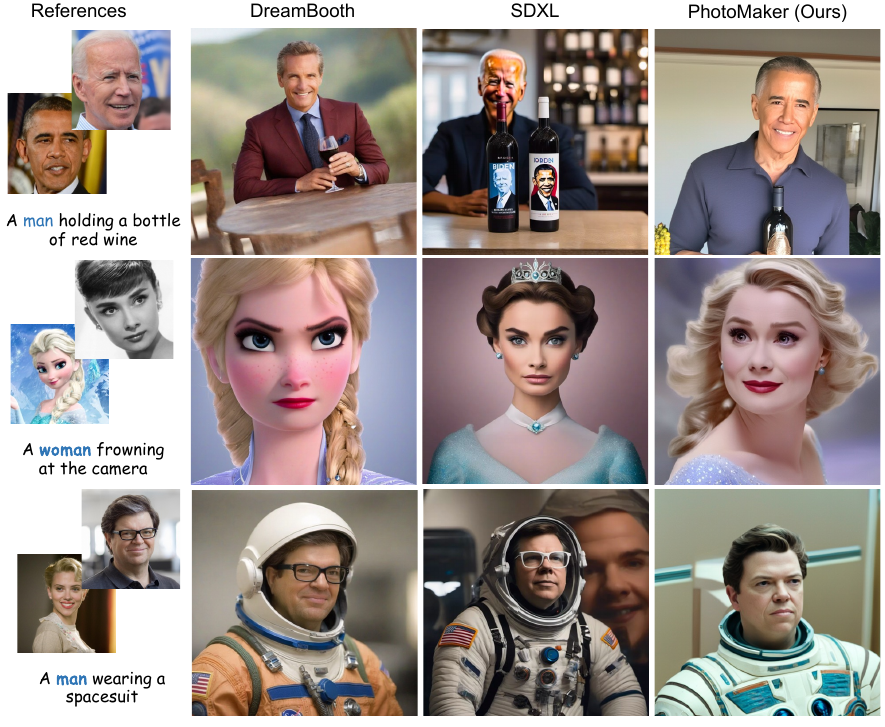}
  \caption{\textbf{Identity mixing}. We are able to generate the image with a new ID while preserving input identity characteristics. 
  We prepare a prompt template \texttt{<original prompt>, with a face blended with <name:A> and <name:B>} for SDXL. (\textit{Zoom-in for the best view})}
  \label{fig:comp_mixing}
  \vspace{-10pt}
\end{figure}

\paragraph{Identity mixing.} 
If the users provide images of different IDs as input, our PhotoMaker can well integrate the characteristics of different IDs to form a new ID.
From \figref{fig:comp_mixing}, we can see that neither DreamBooth nor SDXL can achieve identity mixing.
In contrast,
our method can retain the characteristics of different IDs well on the generated new ID, regardless of whether the input is an anime IP or a real person, and regardless of gender.
Besides, we can control the proportion of this ID in the new generated ID  by controlling the corresponding ID input quantity or prompt weighting. 
We show this ability in \figref{fig:idmix_image}-\ref{fig:idmix_promptweight} in the appendix.

\paragraph{Stylization.} 
In \figref{fig:stylization_main}, we demonstrate the stylization capabilities of our method. 
We can see that, in the generated images, our PhotoMaker not only maintains good ID fidelity but also effectively exhibits the style information of the input prompt. 
This reveals the potential for our method to drive more applications.
Additional results are shown in the \figref{fig:stylization} within the appendix.

\begin{figure}
  \centering  \includegraphics[width=0.48\textwidth]{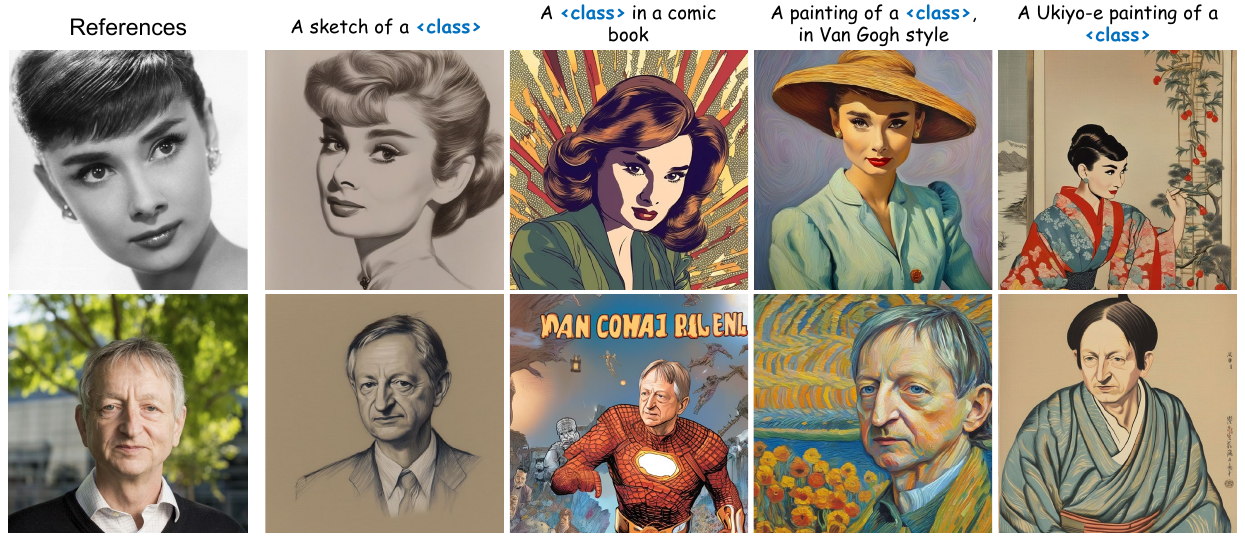}
  \caption{\textbf{The stylization results of our PhotoMaker.}
  The symbol \textbf{{\color{blue} \texttt{<class>}}} denotes it will be replaced by \texttt{man} or \texttt{woman} accordingly.
 \textit{(Zoom-in for the best view)}}
  \label{fig:stylization_main}
\vspace{-10pt}
\end{figure}

\begin{figure*}
  \centering  \includegraphics[width=1.0\textwidth]{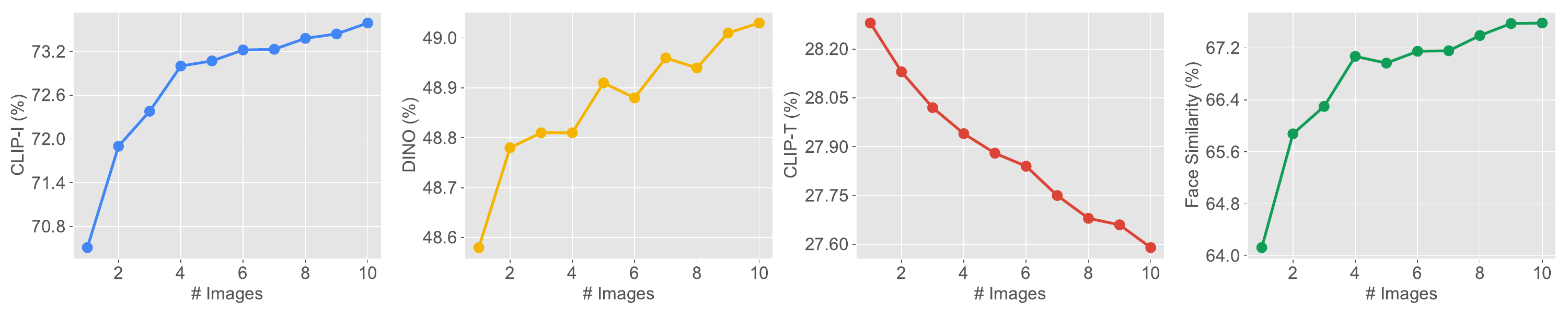}
  \vspace{-30pt}
  \tablestyle{5pt}{1.15}
    \begin{center}
    \begin{tabular}{>{\centering\arraybackslash}p{0.26\textwidth}>{\centering\arraybackslash}p{0.21\textwidth}>{\centering\arraybackslash}p{0.24\textwidth}>{\centering\arraybackslash}p{0.22\textwidth}}
    (a) & (b) & (c) & (d) \\
    \end{tabular}
    \end{center}
  \vspace{-15pt}
  \caption{\textbf{The impact of the number of input ID images on (a) CLIP-I, (b) DINO, (c) CLIP-T, and (d) Face Similarity, respectively.}}
  \label{fig:ablation_num_stacked}
  
\end{figure*}

\begin{table*}
    \centering
    \tablestyle{5pt}{1.15}
    \begin{subtable}[h]{0.48\linewidth}
    \centering
    \begin{tabular}{lcccc}
    \toprule
         & CLIP-T$\uparrow$& DINO$\uparrow$&  Face Sim.$\uparrow$  &Face Div.$\uparrow$\\
    \midrule
         Average& \sota{28.7}& 47.0& 48.8 & \sota{56.3}\\
         Linear& 28.6 & 47.3& 48.1 &54.6\\
         Stacked& 28.0 & \sota{49.5}& \sota{53.6} &55.0\\
    \bottomrule
    \end{tabular}
    \caption{\textbf{Embedding composing choices}.}
    \label{tab:ablation_embed}
     \end{subtable}
     \begin{subtable}[h]{0.48\linewidth}
     \tablestyle{5pt}{1.15}
    \begin{tabular}{lcccc}
    \toprule
         & CLIP-T$\uparrow$&  DINO$\uparrow$&  Face Sim.$\uparrow$&  Face Div.$\uparrow$ \\
    \midrule
         Single embed& 27.9& 50.3& 50.5& \sota{56.1}\\
         Single image& 27.3& \sota{50.3}& \sota{60.4}& 51.7\\
         Ours& \sota{28.0}& 49.5& 53.6& 55.0\\
    \bottomrule
    \end{tabular}
    
    \caption{\textbf{Training data sampling strategy.}}
    \label{tab:ablation_training}
     \end{subtable}
    \vspace{-6pt}
     \caption{\textbf{Ablation studies for the proposed PhotoMaker.} The best results are marked in \sota{bold}.
}
\vspace{-12pt}
\end{table*}

\subsection{Ablation study}
We shortened the total number of training iterations by eight times to conduct ablation studies for each variant.

\begin{figure}
  \centering  \includegraphics[width=0.48\textwidth]{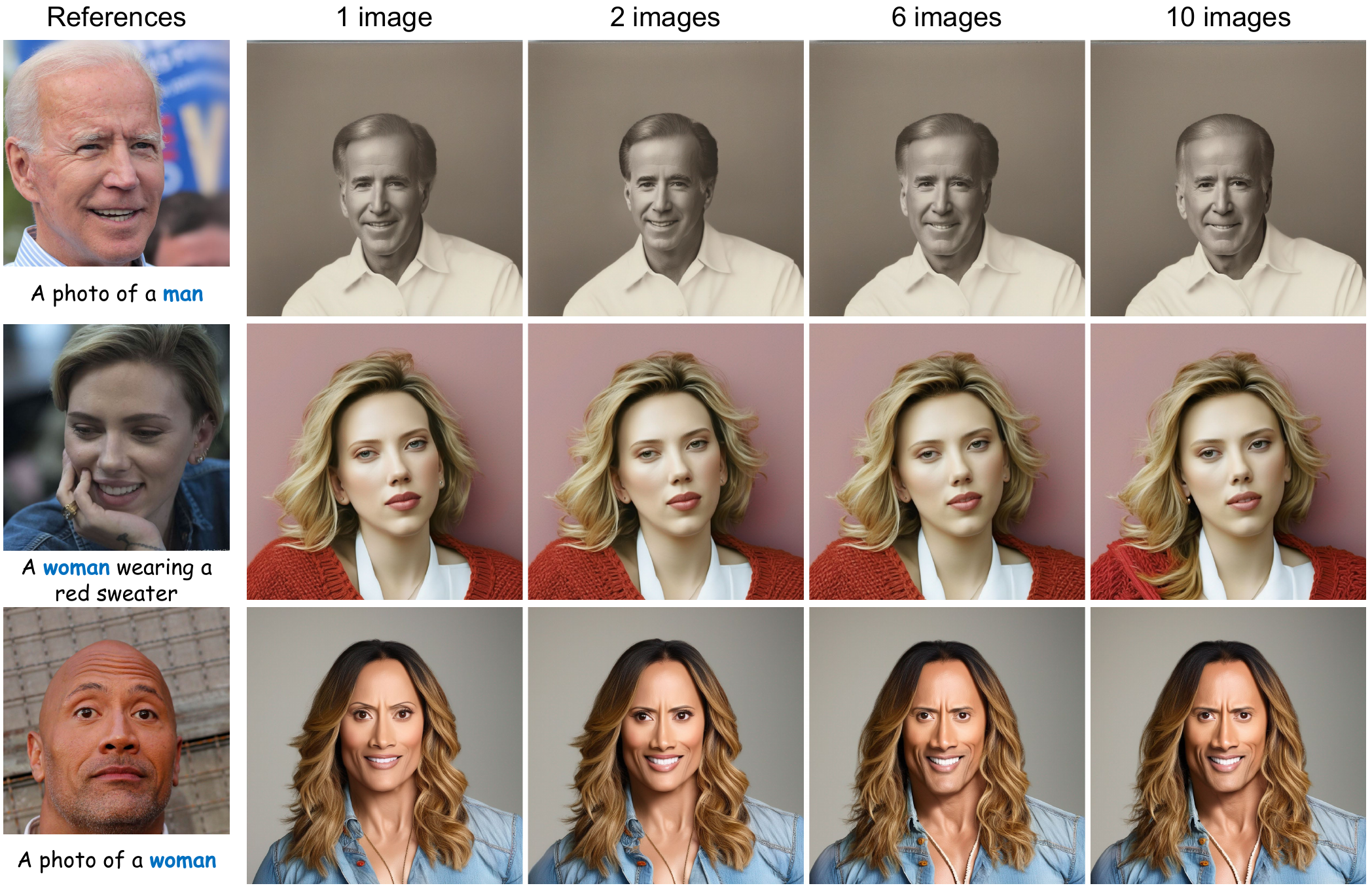}
  \caption{\textbf{The impact of varying the quantity of input images on the generation results.}
 It can be observed that the fidelity of the ID increases with the quantity of input images.}
  \label{fig:num_input_main}
  
\end{figure}

\paragraph{The influence about the number of input ID images.}
We explore the impact that forming the proposed stacked ID embedding through feeding different numbers of ID images.
In \figref{fig:ablation_num_stacked}, we visualize this impact across different metrics.
We conclude that using more images to form a stacked ID embedding can improve the metrics related to ID fidelity. 
This improvement is particularly noticeable when the number of input images is increased from one to two.
Upon the input of an increasing number of ID images, the growth rate of the values in the ID-related metrics significantly decelerates.
Additionally, we observe a linear decline on the CLIP-T metric.
This indicates there may exist a trade-off between text controllability and ID fidelity.
From \figref{fig:num_input_main}, we see that increasing the number of input images enhances the similarity of the ID.
Therefore, the more ID images to form the stacked ID embedding can help the model perceive more comprehensive ID information, and then more accurately represent the ID to generate images.
Besides, as shown by the Dwayne Johnson example, the gender editing capability decreases, and the model is more prone to generate images of the original ID's gender.

\paragraph{The choices of composing multiple embeddings.}
We explore three ways to compose the ID embedding, including averaging the image embeddings, adaptively projecting embeddings through a linear layer, and our stacking way.
From \tabref{tab:ablation_embed}, we see the stacking way has the highest ID fidelity while ensuring a diversity of generated faces, demonstrating its effectiveness.
Besides, such a way offers greater flexibility than others, including accepting any number of images and better controlling the mixing process of different IDs.

\paragraph{The benefits from multiple embeddings during training.}
We explore two other training data sampling strategies to demonstrate that it is necessary to input multiple images with variations during training.
The first is to choose only one image, which can be different from the target image, to form the ID embedding (see ``single embed" in \tabref{tab:ablation_training}). 
Our multiple embedding way has advantages in ID fidelity.
The second sampling strategy is to regard the target image as the input ID image (to simulate the training way of most embedding-based methods). 
We generate multiple images based on this image with different data augmentation methods and extract corresponding multiple embeddings. 
In \tabref{tab:ablation_training}, as the model can easily remember other irrelevant characteristics of the input image, the generated facial area lacks sufficient changes (low diversity).

%% file: sec/5_conclusion.tex
\section{Conclusion}

We have presented \textit{PhotoMaker}, an efficient personalized text-to-image generation method that focuses on generating realistic human photos.
Our method leverages a simple yet effective representation, stacked ID embedding, for better preserving ID information.
Experimental results have demonstrated that our PhotoMaker, compared to other methods, can simultaneously satisfy high-quality and diverse generation capabilities, promising editability, high inference efficiency, and strong ID fidelity.
Besides, we also have found that our method can empower many interesting applications that previous methods are hard to achieve, such as changing age or gender, bringing persons from old photos or artworks back to reality, and identity mixing.

%% file: sec/X_suppl.tex
\clearpage
\setcounter{page}{1}
\setcounter{section}{0}
\makearxivsupplementary

\renewcommand\thesection{\Alph{section}}
\section{Dataset Details}

\paragraph{Training dataset.}
Based on \secref{sec:data_pipeline} in the main paper,
following a sequence of filtering steps,
the number of images in our constructed dataset is about 112K.
They are classified by about 13,000 ID names.
Each image is accompanied by a mask for the corresponding ID and an annotated caption.

\paragraph{Evaluation dataset.}
The \textit{image dataset} used for evaluation comprises manually selected additional IDs and a portion of MyStyle~\cite{nitzan2022mystyle} data. 
For each ID name, we have four images that serve as input data for comparative methods and for the final metric evaluation (\ie, DINO~\cite{caron2021emerging}, CLIP-I~\cite{gal2022image}, and Face Sim.~\cite{deng2019arcface}).
For single-embedding methods (\ie, FastComposer~\cite{xiao2023fastcomposer} and IPAdapter~\cite{ye2023ip}), we randomly select one image from each ID group as input. 
Note that the ID names exist in the training image set, utilized for the training of our method, and the test image set, do not exhibit any overlap.
We list ID names for evaluation in ~\tabref{tab:evaluate_ids}.
For \textit{text prompts} used for evaluation,  we consider six factors: clothing, accessories, actions, expressions, views, and background, which make up 40 prompts that are listed in the \tabref{tab:evaluate_prompts}.

\begin{table}
    \centering
    \begin{tabular}{ll}
    \toprule
        \multicolumn{2}{c}{Evaluation IDs}\\
    \midrule
   \Circled{1} Alan Turing
 &\Circled{14} Kamala Harris\\
\Circled{2} Albert Einstein &\Circled{15} Marilyn Monroe\\
      \Circled{3}  Anne Hathaway &\Circled{16} Mark Zuckerberg\\
\Circled{4} Audrey Hepburn &\Circled{17} Michelle Obama\\
        \Circled{5} Barack Obama&\Circled{18} Oprah Winfrey\\
\Circled{6} Bill Gates&\Circled{19} Renée Zellweger\\
\Circled{7} Donald Trump&\Circled{20} Scarlett Johansson\\
        \Circled{8} Dwayne Johnson&\Circled{21} Taylor Swift\\
\Circled{9} Elon Musk&\Circled{22} Thomas Edison\\
\Circled{10} Fei-Fei Li&\Circled{23} Vladimir Putin\\
\Circled{11} Geoffrey Hinton&\Circled{24} Woody Allen\\
\Circled{12} Jeff Bezos&\Circled{25} Yann LeCun\\
        \Circled{13} Joe Biden&\\
    \bottomrule
    \end{tabular}
    \caption{\textbf{ID names used for evaluation.} For each name, we collect four images totally.}
    \label{tab:evaluate_ids}
\end{table}

\begin{table*}
    \centering
    \tablestyle{5pt}{1.15}
    \begin{subtable}[h]{0.48\linewidth}
    \centering
    \begin{tabular}{c|p{0.7\linewidth}}
    \toprule
    Category  & Prompt \\
    \midrule
    General  & a photo of a \texttt{<class word>} \\
    \hline
    \multirow{5}{*}{Clothing}   &  a \texttt{<class word>} wearing a Superman outfit \\
         & a \texttt{<class word>} wearing a spacesuit \\
         & a \texttt{<class word>} wearing a red sweater \\
         & a \texttt{<class word>} wearing a purple wizard outfit \\
         & a \texttt{<class word>} wearing a blue hoodie
 \\
\hline
 \multirow{10}{*}{Accessory}&  a \texttt{<class word>} wearing headphones\\
         &  a \texttt{<class word>} with red hair\\
         &  a \texttt{<class word>} wearing headphones with red hair\\
         &  a \texttt{<class word>} wearing a Christmas hat\\
         &  a \texttt{<class word>} wearing sunglasses\\
         &  a \texttt{<class word>} wearing sunglasses and necklace\\
         &  a \texttt{<class word>} wearing a blue cap\\
         &  a \texttt{<class word>} wearing a doctoral cap\\
         &  a \texttt{<class word>} with white hair, wearing glasses\\
\hline
 \multirow{10}{*}{Action}&  a \texttt{<class word>} in a helmet and vest riding a motorcycle\\
         &  a \texttt{<class word>} holding a bottle of red wine\\
         & a \texttt{<class word>} driving a bus in the desert\\
         &  a \texttt{<class word>} playing basketball\\
         &  a \texttt{<class word>} playing the violin\\
         &  a \texttt{<class word>} piloting a spaceship\\
         &  a \texttt{<class word>} riding a horse\\
         &  a \texttt{<class word>} coding in front of a computer\\
         &  a \texttt{<class word>} playing the guitar\\
    \bottomrule
    \end{tabular}
    \caption{}
    \end{subtable}
    \begin{subtable}[h]{0.48\linewidth}
    \centering
        \begin{tabular}{c|p{0.7\linewidth}}
        \toprule
            Category  & Prompt \\
        \midrule
         \multirow{6}{*}{Expression}& a \texttt{<class word>} laughing on the lawn\\
         &  a \texttt{<class word>} frowning at the camera\\
         &  a \texttt{<class word>} happily smiling, looking at the camera\\
         &  a \texttt{<class word>} crying disappointedly, with tears flowing\\
         &  a \texttt{<class word>} wearing sunglasses\\
         \hline
          \multirow{14}{*}{View}& a \texttt{<class word>} playing the guitar in the view of left side\\
         &  a \texttt{<class word>} holding a bottle of red wine, upper body\\
         &  a \texttt{<class word>} wearing sunglasses and necklace, close-up, in the view of right side\\
         &  a \texttt{<class word>} riding a horse, in the view of the top\\
         &  a \texttt{<class word>} wearing a doctoral cap, upper body, with the left side of the face facing the camera\\
         &  a \texttt{<class word>} crying disappointedly, with tears flowing, with left side of the face facing the camera\\
         \hline
         \multirow{8}{*}{Background}& a \texttt{<class word>} sitting in front of the camera, with a beautiful purple sunset at the beach in the background\\
         &  a \texttt{<class word>} swimming in the pool\\
         &  a \texttt{<class word>} climbing a mountain\\
         &  a \texttt{<class word>} skiing on the snowy mountain\\
         &  a \texttt{<class word>} in the snow\\
         &  a \texttt{<class word>} in space wearing a spacesuit\\
        \bottomrule
        \end{tabular}
    \caption{}
    \end{subtable}
    \caption{\textbf{Evaluation text prompts categorized by (a) general setting, clothing, accessory, action, (b) expression, view, and background.} The \texttt{class word} will be replaced with \texttt{man}, \texttt{woman}, \texttt{boy}, etc. For each ID and each prompt, we randomly generated four images for evaluation.}
    \label{tab:evaluate_prompts}
\end{table*}

\section{User Study}
\label{sec:user_study}

\begin{figure}
  \centering  \includegraphics[width=0.48\textwidth]{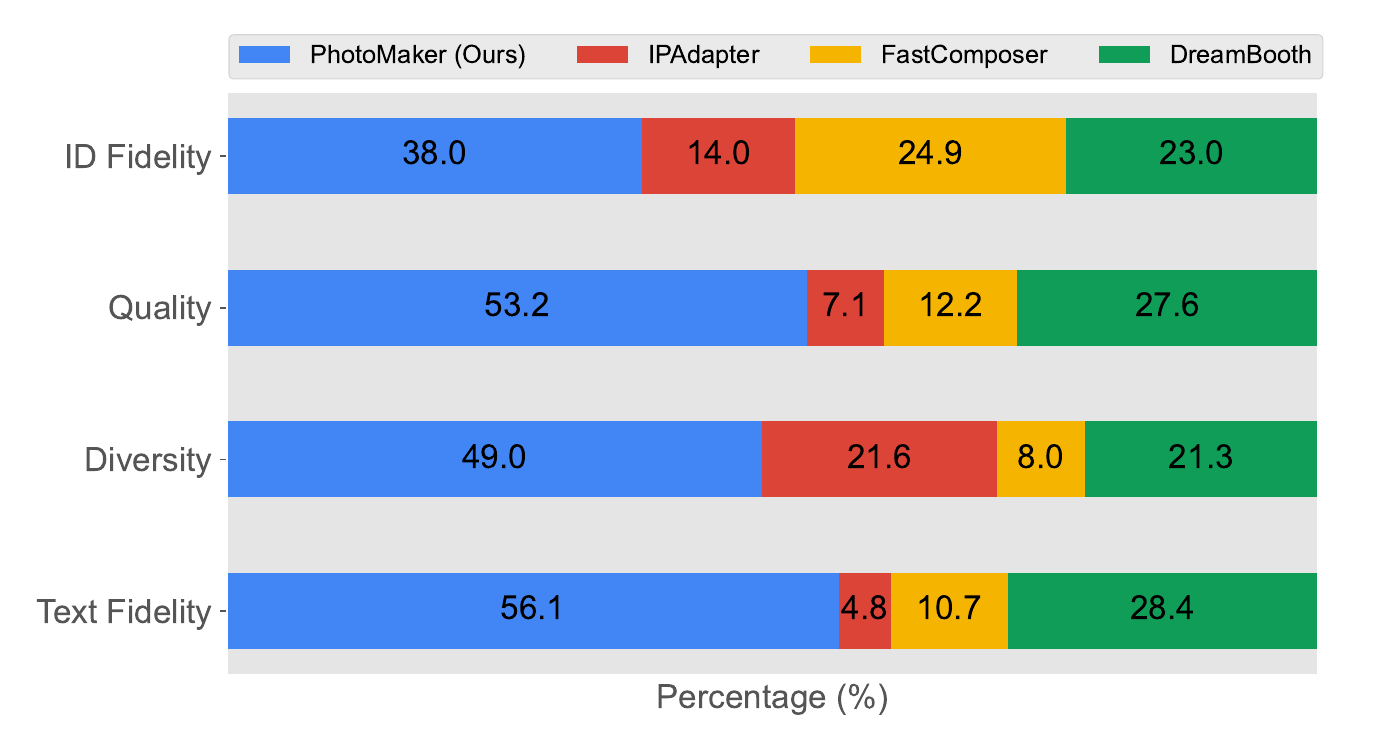}
  \caption{\textbf{User preferences on ID fidelity, generation quality, face diversity, and text fidelity for different methods.}
  For ease of illustration, we visualize the proportion of total votes that each method has received.
  Our PhotoMaker occupies the most significant proportion in these four dimensions.}
  \label{fig:user_study}
  
\end{figure}

In this section, we conduct a user study to make a more comprehensive comparison.
The comparative methods we have selected include DreamBooth~\cite{ruiz2022dreambooth}, FastComposer~\cite{xiao2023fastcomposer}, and IPAdapter~\cite{ye2023ip}.
We use SDXL~\cite{podell2023sdxl} as the base model for both DreamBooth and IPAdapter because of their open-sourced implementations.
We display 20 text-image pairs for each user.
Each of them includes a reference image of the input ID and corresponding text prompt.
We have four randomly generated images of each method for each text-image pair. 
Each user is requested to answer four questions for these 20 sets of results:
1) Which method is \textit{most similar} to the input person's \textit{identity}? 
2) Which method produces the \textit{highest quality} generated images? 
3) Which method generates \textit{the most diverse} facial area in the images?  
4) Which method generates images that \textit{best match} the input \textit{text prompt}?
We have anonymized the names of all methods and randomized the order of methods in each set of responses.
We had a total of 40 candidates participating in our user study, and we received 3,200 valid votes.
The results are shown in \figref{fig:user_study}.

We find that our PhotoMaker has advantages in terms of ID fidelity, generation quality, diversity, and text fidelity, especially the latter three.
In addition, we found that DreamBooth is the second-best algorithm in balancing these four evaluation dimensions, which may explain why it was more prevalent than the embedding-based methods
in the past.
At the same time, IPAdapter shows a significant disadvantage in terms of generated image quality and text consistency, as it focuses more on image embedding during the training phase.
FastComposer has a clear shortcoming in the diversity of the facial region for their single-embedding training pipeline.
The above results are generally consistent with \tabref{tab:comp_recontext} in the main paper, except for the discrepancy in the CLIP-T metric. 
This could be due to a preference for selecting images that harmonize with the objects appearing in the text when manually choosing the most text-compatible images. 
In contrast, the CLIP-T tends to focus on whether the object appears.
This may demonstrate the limitations of the CLIP-T.
We also provide more visual samples in \figref{fig:supp_comp_recontext_user_study}-\ref{fig:supp_comp_painting} for reference.

\section{More Ablations}

\begin{figure*}
  \centering  \includegraphics[width=0.98\textwidth]{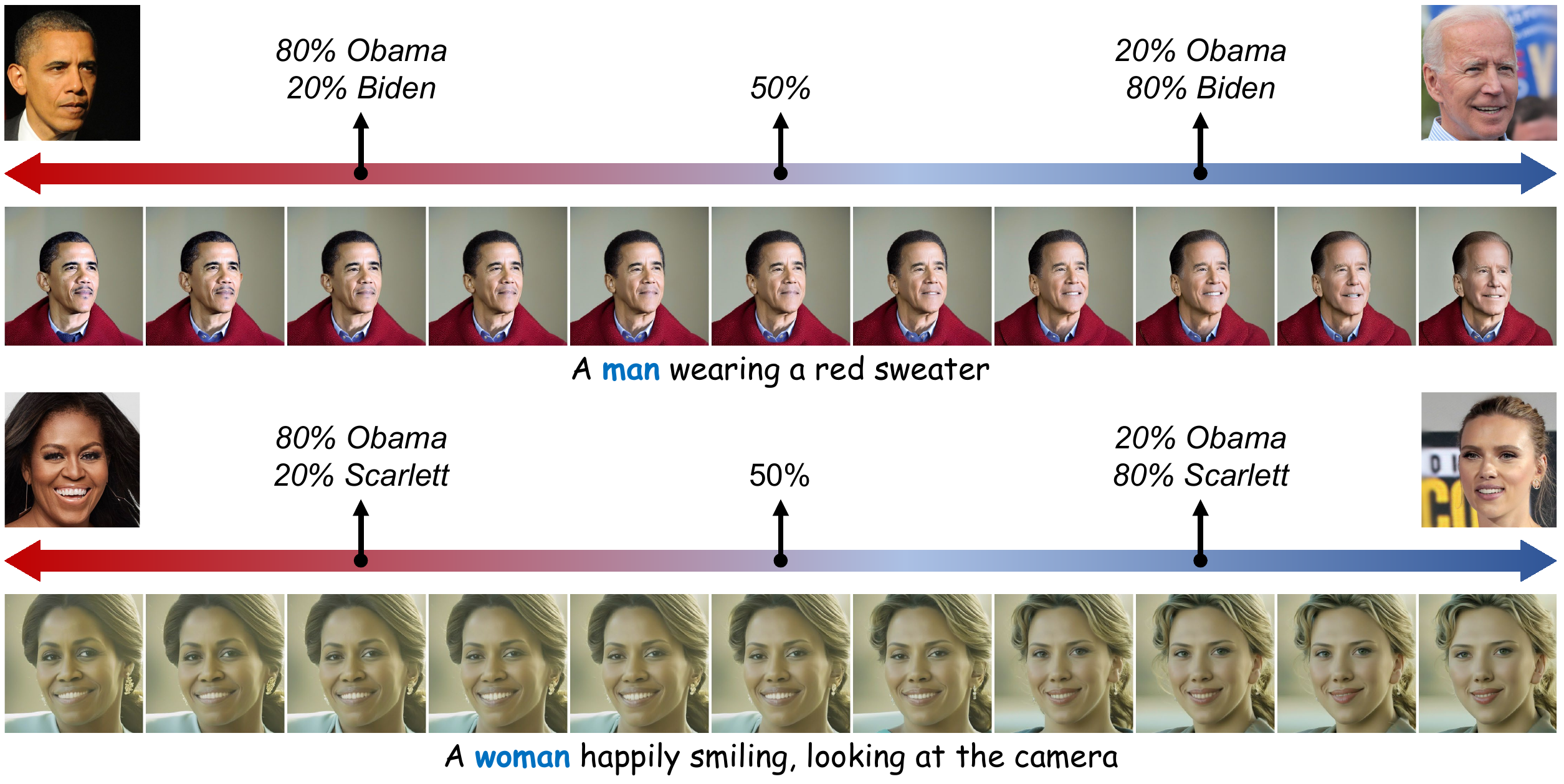}
  \caption{\textbf{The impact of the proportion of images with different IDs in the input sample pool on the generation of new IDs.}
The first row illustrates the transition from Barack Obama to Joe Biden. 
The second row depicts the shift from Michelle Obama to Scarlett Johansson.
To provide a clearer illustration, percentages are used in the figure to denote the proportion of each ID in the input image pool. 
The total number of images contained in the input pool is 10.
\textit{(Zoom-in for the best view)}}
\label{fig:idmix_image}
\end{figure*}

\begin{figure*}
  \centering  \includegraphics[width=0.98\textwidth]{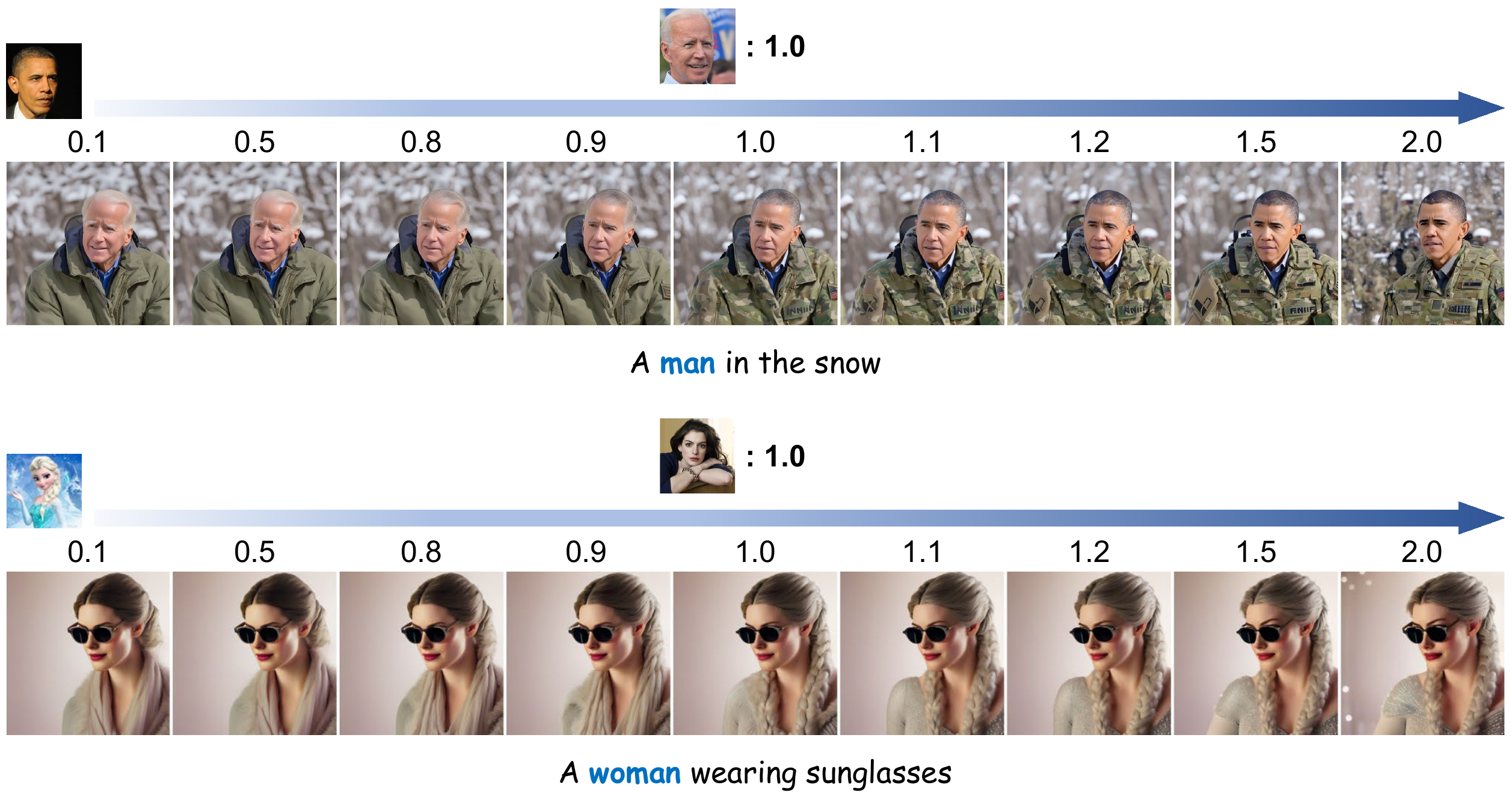}
  \caption{\textbf{The impact of prompt weighting on the generation of new IDs.}
The first row illustrates a blend of Barack Obama and Joe Biden. 
The first row from left to right represents the progressive increase in the weight of the ID image embedding corresponding to Barack Obama in the image.
The second row illustrates a blend of Elsa (Disney) and Anne Hathaway. 
The weight for Elsa is gradually increased.
 \textit{(Zoom-in for the best view)}}
  \label{fig:idmix_promptweight}
\end{figure*}

\paragraph{Adjusting the ratio during identity mixing.}
For identity mixing, our method can adjust the merge ratio by either controlling the percentage of identity images within the input image pool or through the method of prompt weighting~\cite{hertz2022prompt, prompt_weighting}.
In this way, we can control that the person generated with a new ID is either more closely with or far away from a specific input ID.
\figref{fig:idmix_image} shows how our method customizes a new ID by controlling the proportion of different IDs in the input image pool. 
For a better description, we use a total of 10 images as input in this experiment. 
We can observe a smooth transition of images with the two IDs.
This smooth transition encompasses changes in skin color and age.
Next, we use four images per generated ID to conduct prompt weighting.
The results are shown in \figref{fig:idmix_promptweight}.
We multiply the embedding corresponding to the images related to a specific ID by a coefficient to control its proportion of integration into the new ID.
Compared to the way to control the number of input images,
prompt weighting requires fewer photos to adjust the merge ratio of different IDs, demonstrating its superior usability.
Besides,
the two ways of adjusting the mixing ratio of different IDs both demonstrate the flexibility of our method.

\begin{figure*}
  \centering  \includegraphics[width=0.95\textwidth]{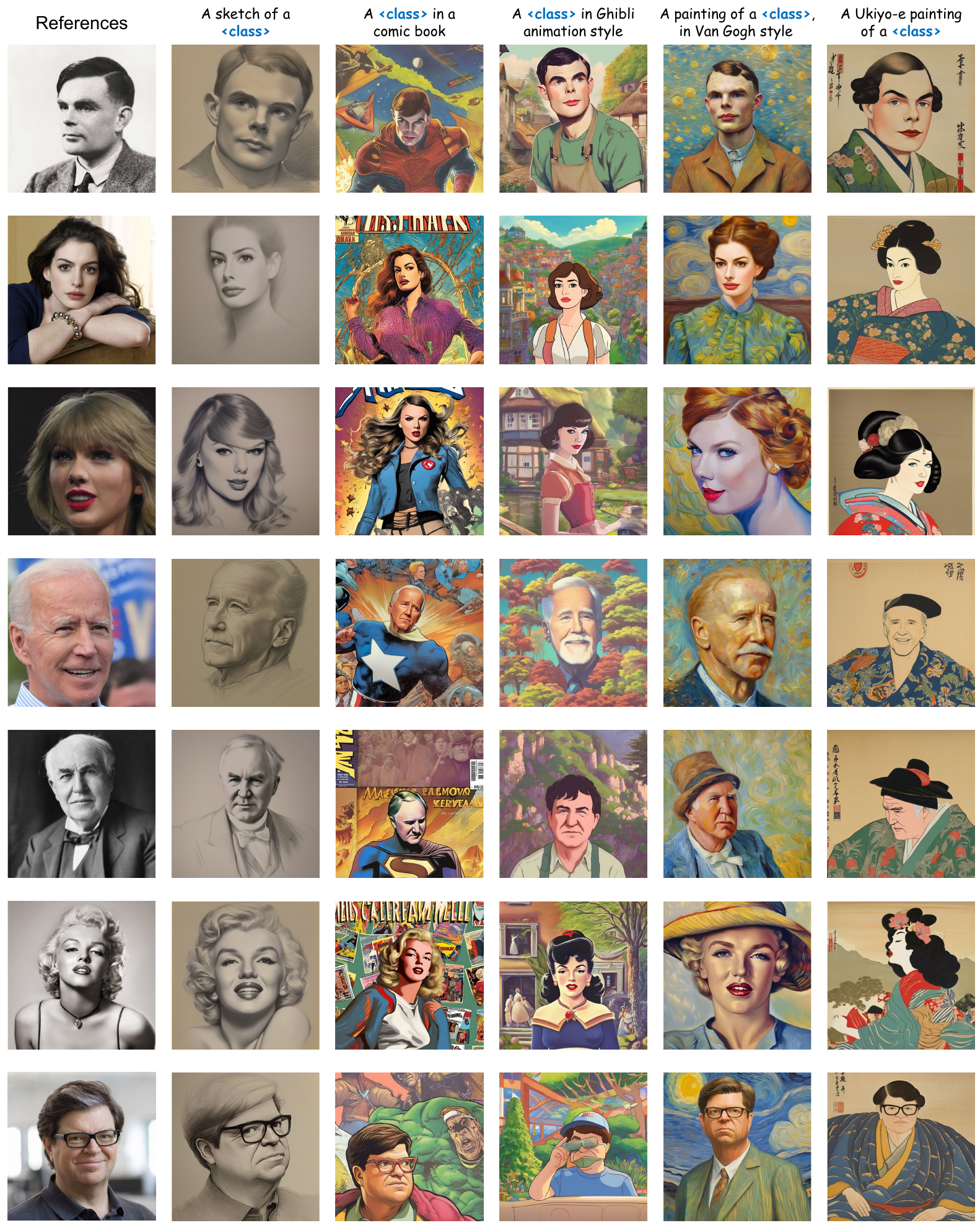}
  \caption{\textbf{The stylization results of our PhotoMaker with different input IDs and different style prompts.}
  Our method can be seamlessly transferred to a variety of styles, concurrently preventing the generation of realistic results.
  The symbol \textbf{{\color{blue} \texttt{<class>}}} denotes it will be replaced by \texttt{man} or \texttt{woman} accordingly.
 \textit{(Zoom-in for the best view)}}
  \label{fig:stylization}
  
\end{figure*}

\section{Stylization Results}
Our method not only possesses the capability to generate realistic human photos, but it also allows for stylization while preserving ID attributes.
This demonstrates the robust generalizability of the proposed method.
We provide the stylization results in \figref{fig:stylization}.

\section{More Visual Results}

\begin{figure}
  \centering  \includegraphics[width=0.48\textwidth]{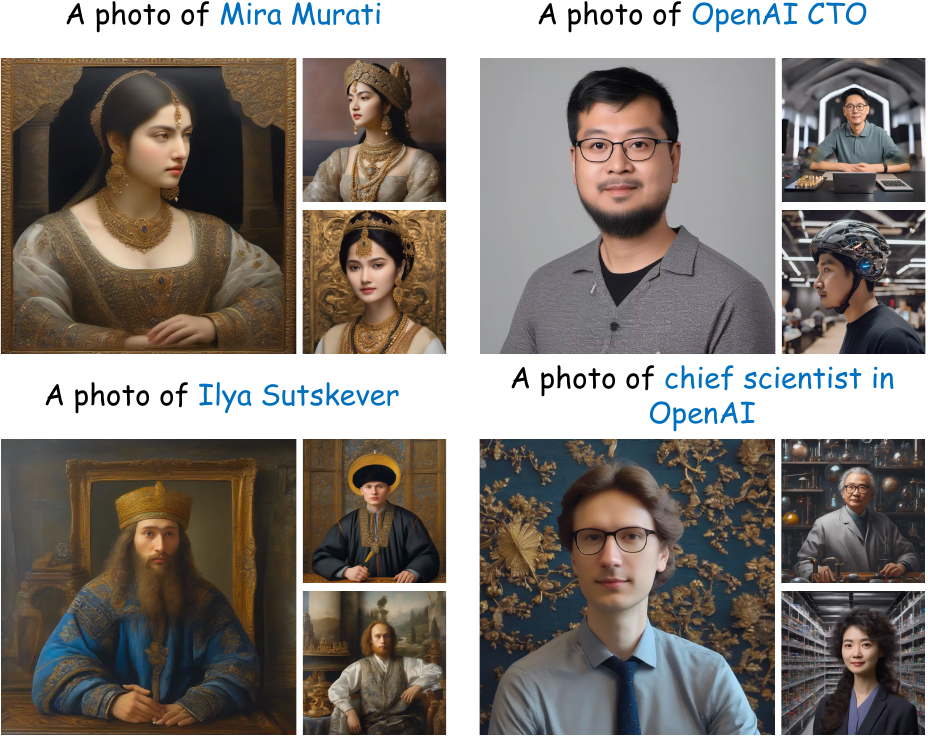}
  \caption{\textbf{Two examples that are unrecognizable by the SDXL.} We replaced two types of text prompts (\eg, name and position) but were unable to prompt the SDXL to generate Mira Murati and Ilya Sutskever.}
  \label{fig:sdxl_failed}
  
\end{figure}

\begin{figure*}
  \centering  \includegraphics[width=0.98\textwidth]{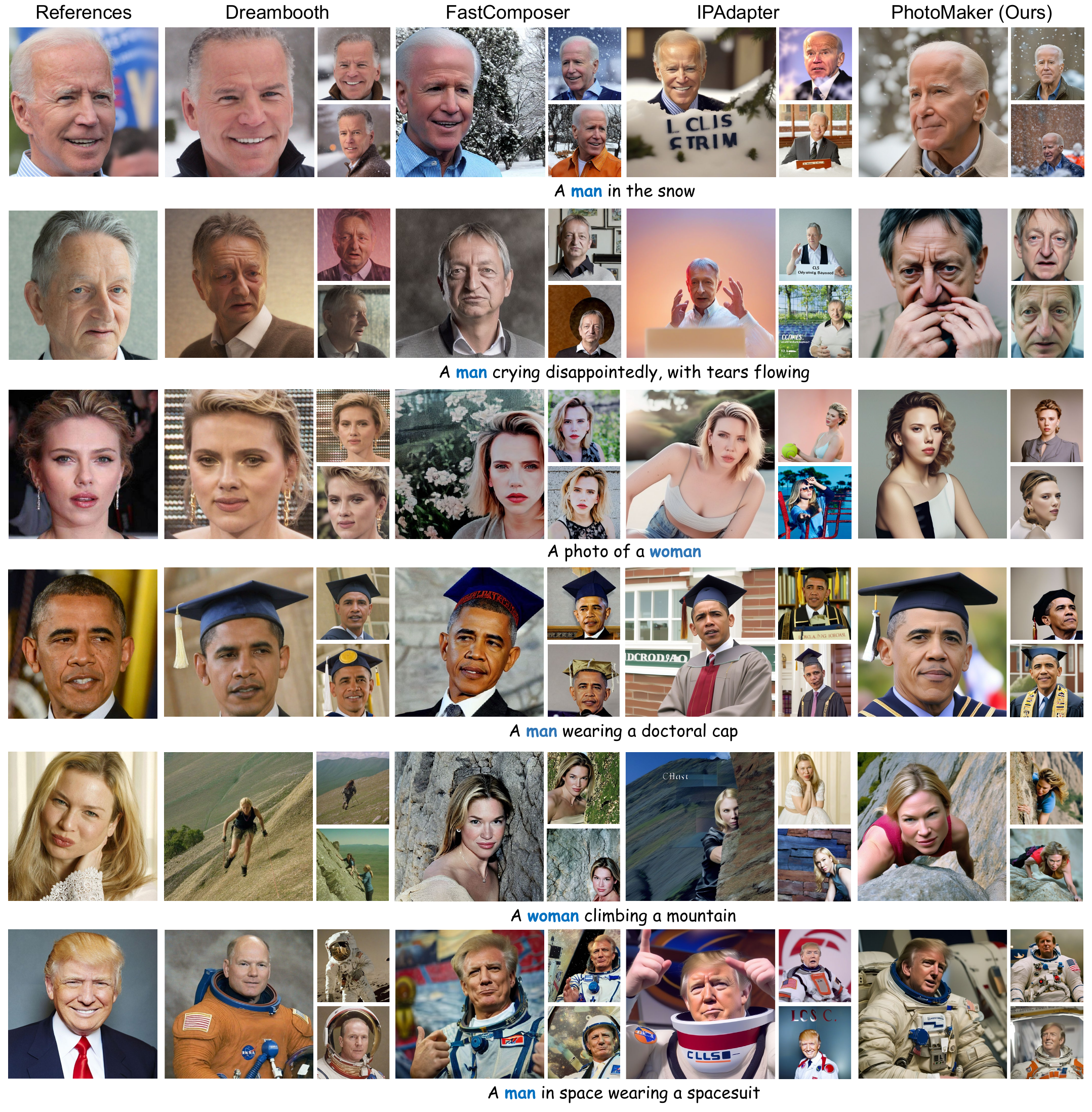}
  \caption{\textbf{More visual examples for recontextualization setting.}
 Our method not only provides high ID fidelity but also retains text editing capabilities.
 We randomly sample three images for each prompt. \textit{(Zoom-in for the best view)}
 }
  \label{fig:supp_comp_recontext_user_study}
\end{figure*}

\begin{figure*}
  \centering  \includegraphics[width=0.98\textwidth]{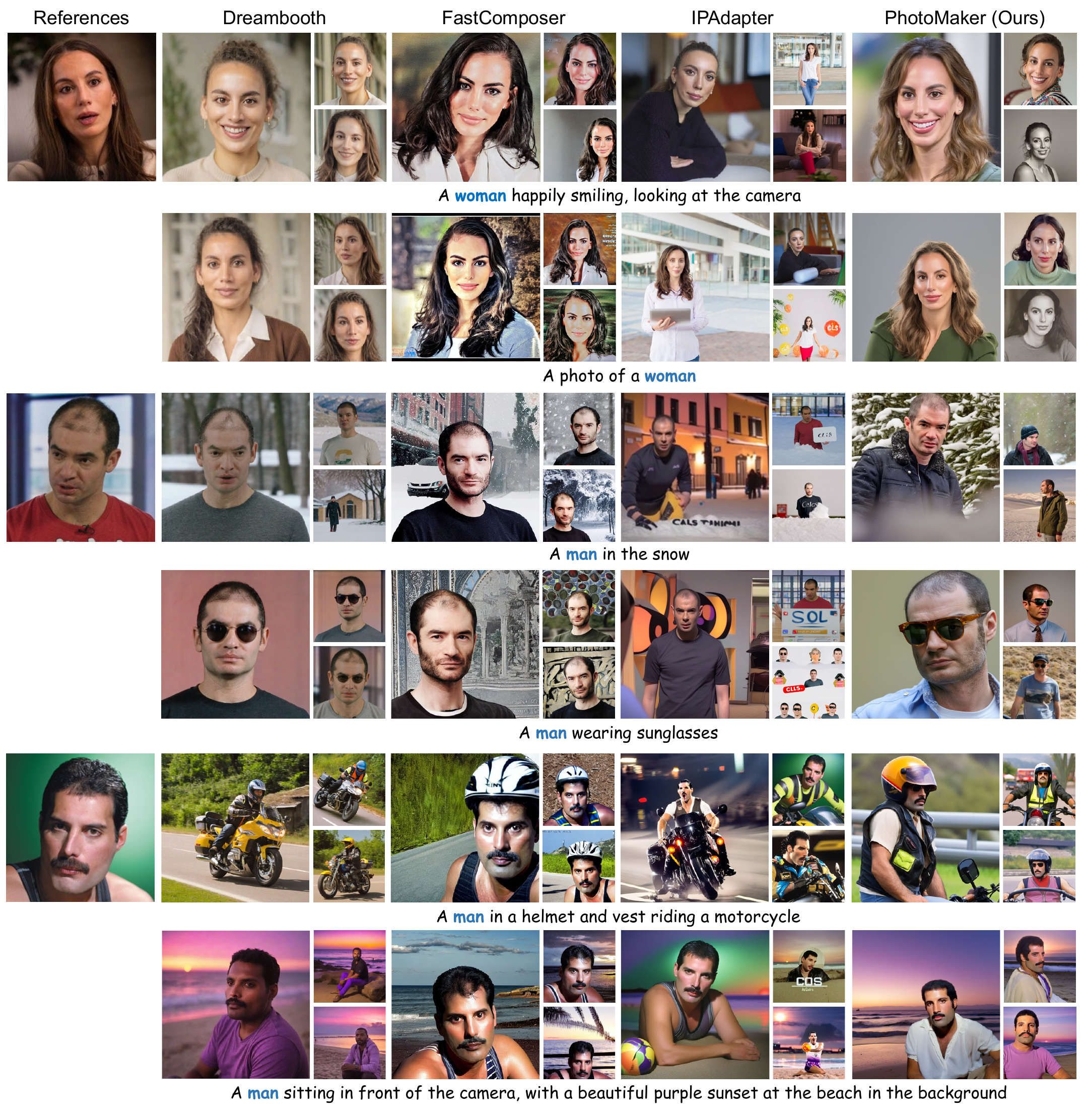}
  \caption{\textbf{More visual examples for recontextualization setting.}
 Our method not only provides high ID fidelity but also retains text editing capabilities.
 We randomly sample three images for each prompt. \textit{(Zoom-in for the best view)
 }
 }
  \label{fig:supp_comp_recontextual}
\end{figure*}

\paragraph{Recontextualization.}
We first provide a more intuitive comparison in \figref{fig:supp_comp_recontext_user_study}.
We compare our PhotoMaker with DreamBooth~\cite{ruiz2022dreambooth}, FastComposer~\cite{xiao2023fastcomposer}, and IPAdapater~\cite{ye2023ip}, for universal recontextualization cases.
Compared to other methods, the results generated by our method can simultaneously satisfy high-quality, strong text controllability, and high ID fidelity.
We then focus on the IDs that SDXL can not generate itself.
We refer to this scenario as the ``non-celebrity" case.
Compared \figref{fig:supp_comp_recontextual} with \figref{fig:sdxl_failed},
our method can successfully generate the corresponding input IDs for this setting.

\begin{figure*}
  \centering  \includegraphics[width=0.98\textwidth]{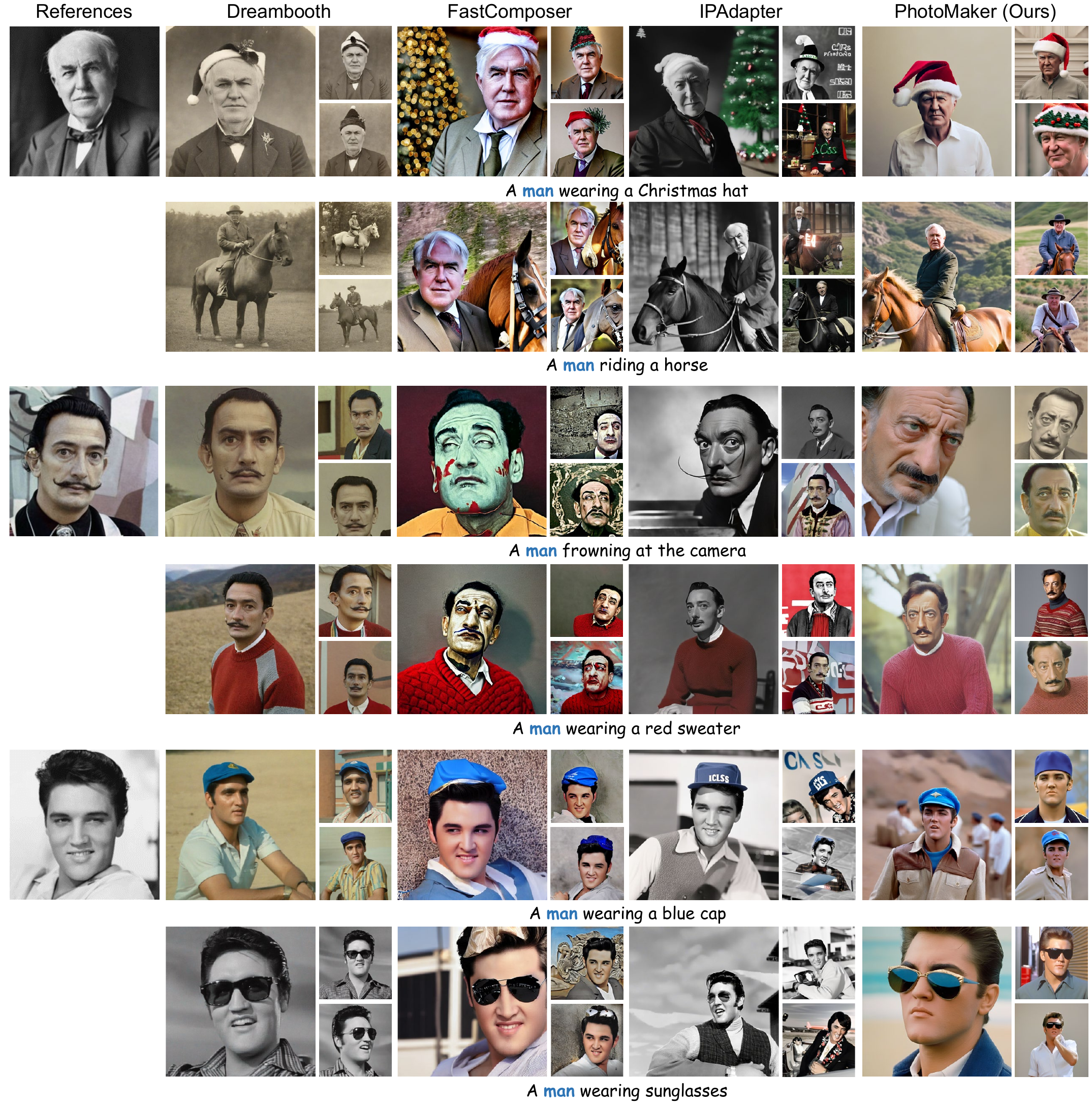}
  \caption{\textbf{More visual examples for bringing person in old-photo back to life.}
 Our method can generate high-quality images.
  We randomly sample three images for each prompt. \textit{(Zoom-in for the best view)}
 }
  \label{fig:supp_comp_oldphoto}
\end{figure*}

\begin{figure*}
  \centering  \includegraphics[width=0.98\textwidth]{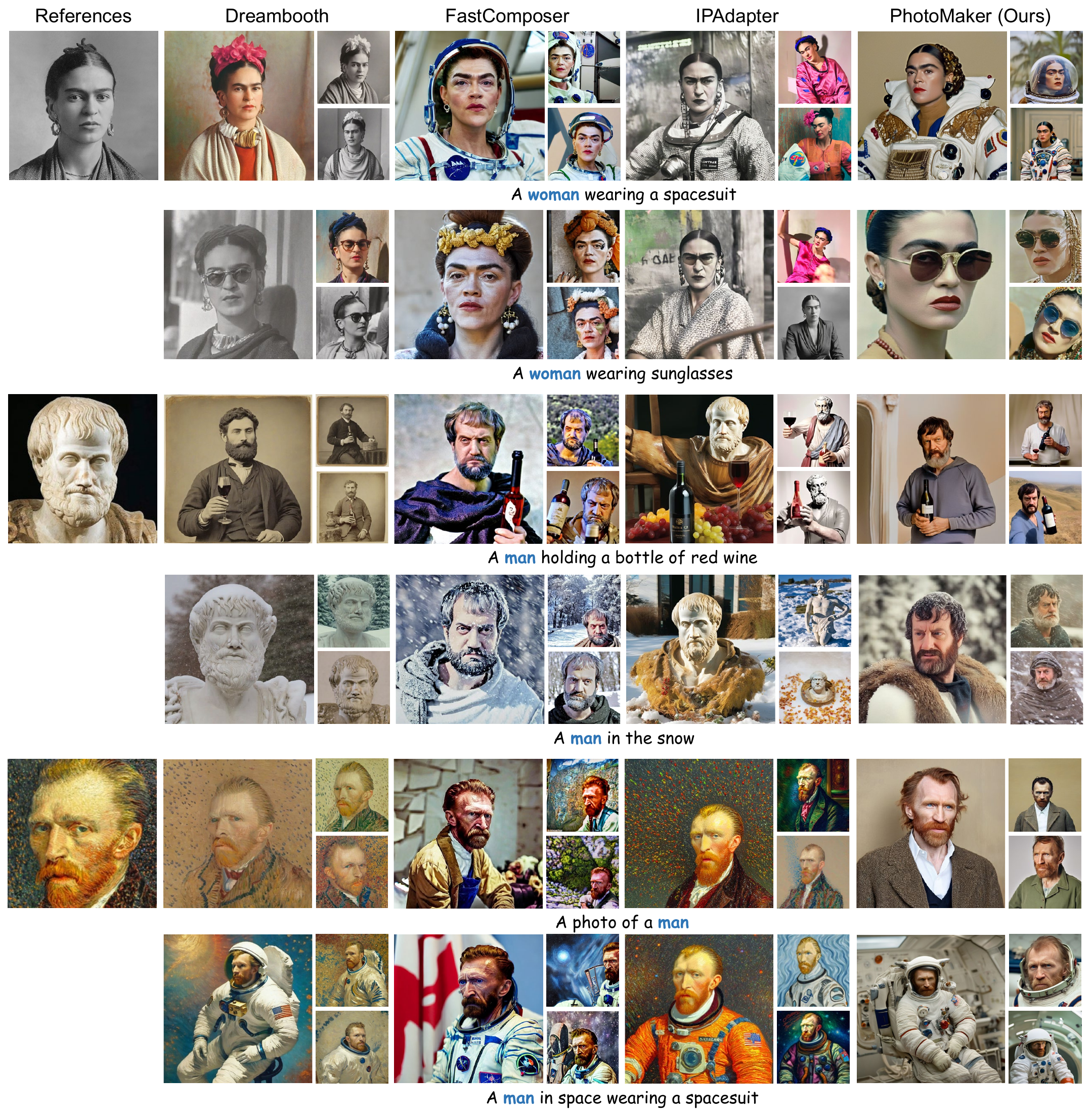}
  \caption{\textbf{More visual examples for bringing person in artworks back to life.}
 Our PhotoMaker can generate photo-realistic images while other methods are hard to achieve.
We randomly sample three images for each prompt. \textit{(Zoom-in for the best view)}
 }
  \label{fig:supp_comp_painting}
\end{figure*}

\paragraph{Bringing person in artwork/old photo into reality.}
\figref{fig:supp_comp_oldphoto}-\ref{fig:supp_comp_painting} demonstrate the ability of our method to bring past celebrities back to reality.
It is worth noticing that 
our method can generate photo-realistic images from IDs in statues and oil paintings.
Achieving this is quite challenging for the other methods we have compared.

\begin{figure*}
  \centering  \includegraphics[width=0.9\textwidth]{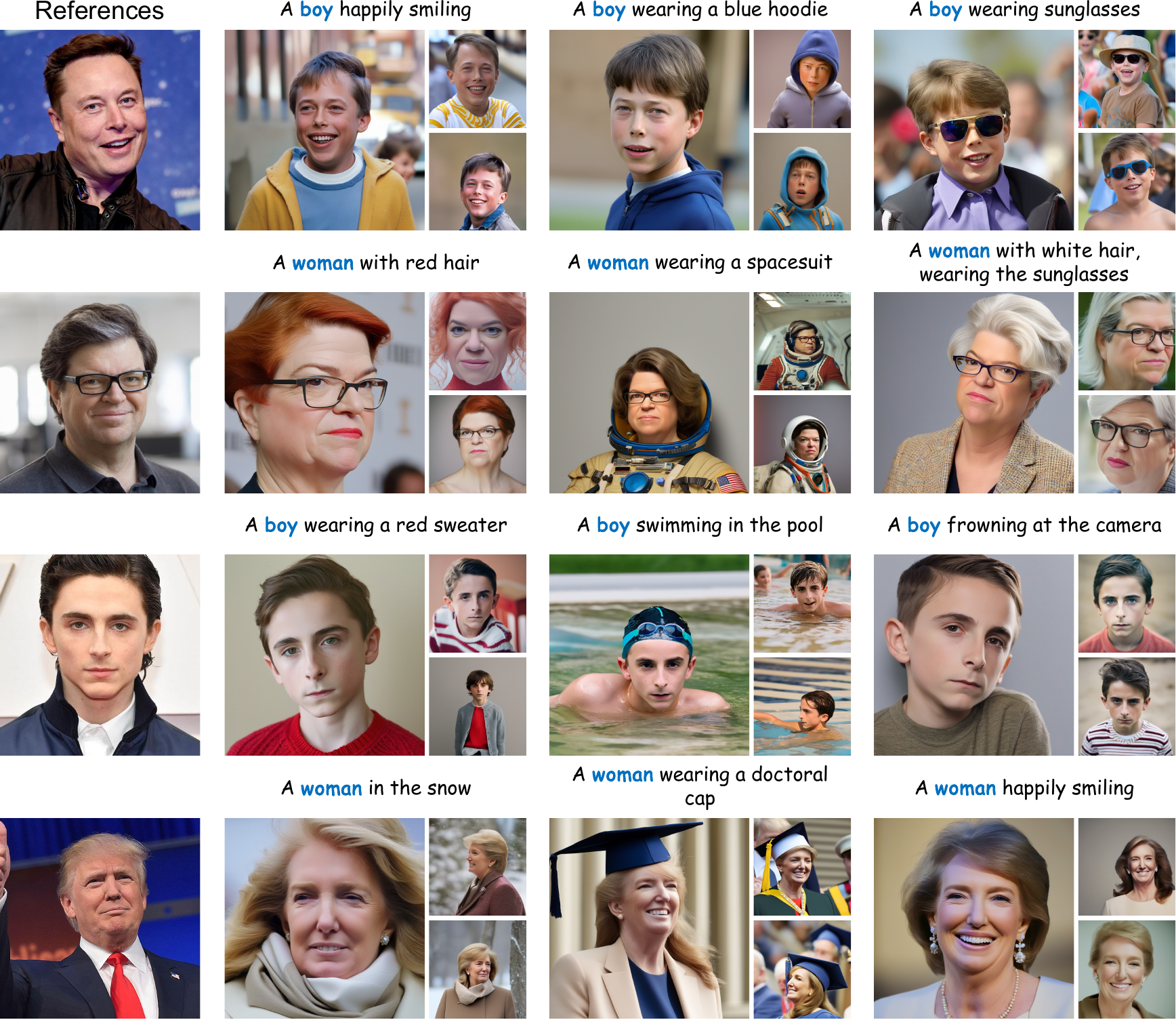}
  \caption{\textbf{More visual examples for changing age or gender for each ID.}
 Our PhotoMaker, when modifying the gender and age of the input ID, effectively retains the characteristics of the face ID and allows for textual manipulation.
   We randomly sample three images for each prompt. \textit{(Zoom-in for the best view)}
 }
  \label{fig:comp_agegender}
\end{figure*}

\begin{figure*}
  \centering  \includegraphics[width=0.9\textwidth]{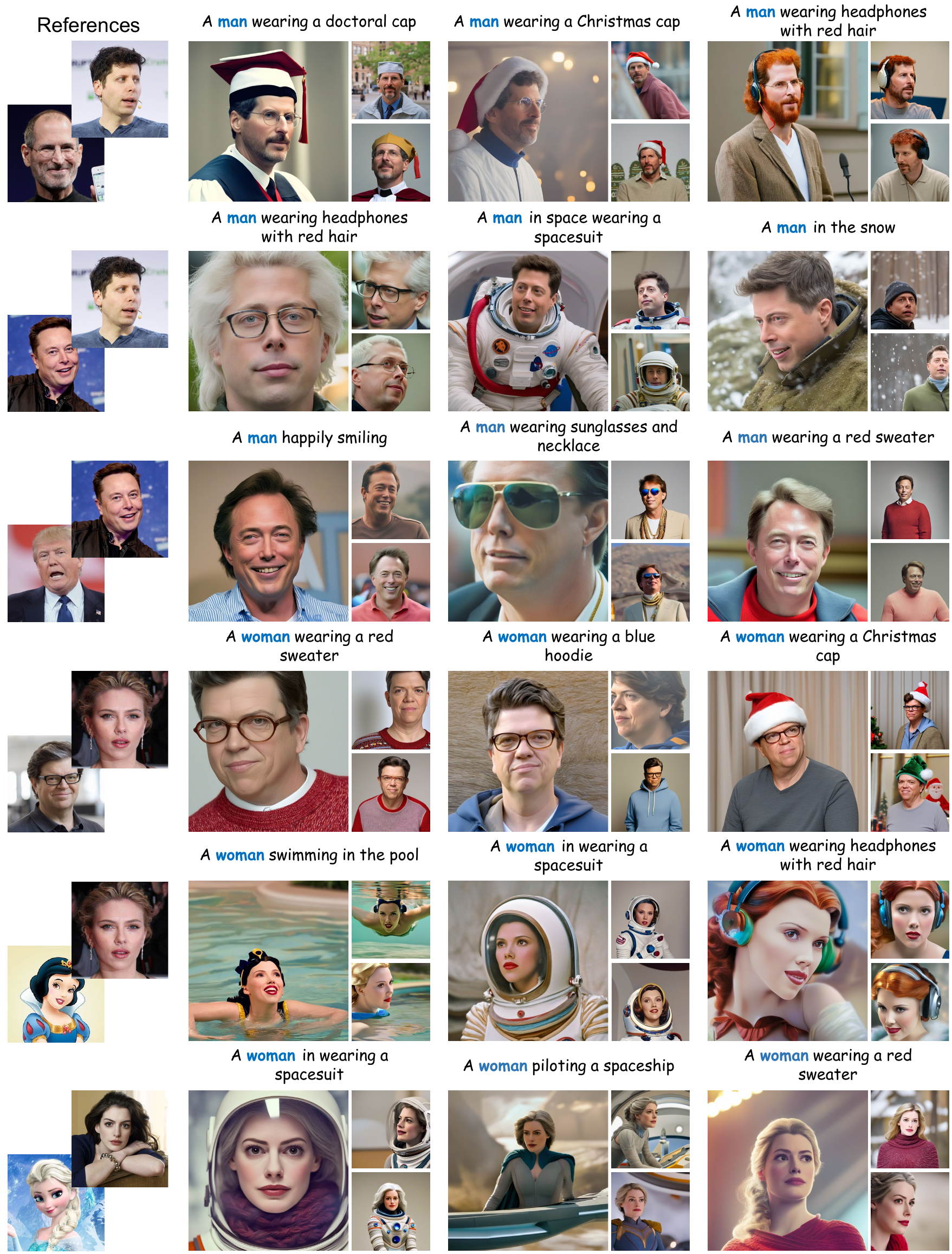}
  \caption{\textbf{More visual results for identity mixing applications.}
  Our PhotoMaker can maintain the characteristics of both input IDs in the new generated ID image, while providing high-quality and text-compatible generation results.
    We randomly sample three images for each prompt. \textit{(Zoom-in for the best view)}
  }
  \label{fig:comp_idmix}
\end{figure*}

\paragraph{Changing age or gender.}
We provide more visual results for changing age or gender in \figref{fig:comp_agegender}.
As mentioned in the main paper, 
we only need to change the class word when we conduct such an application.
In the generated ID images changed in terms of age or gender, our method can well preserve the characteristics in the original ID.

\paragraph{Identity mixing.}
We provide more visual results for identity mixing application in \figref{fig:comp_idmix}.
Benefiting from our stacked ID embedding, our method can effectively blend the characteristics of different IDs to form a new ID. Subsequently, we can generate text controlled based on this new ID.
Additionally, our method provides great flexibility during the identity mixing, as can be seen in \figref{fig:idmix_image}-\ref{fig:idmix_promptweight}.
More importantly, we have explored in the main paper that existing methods struggle to achieve this application.
Conversely, our PhotoMaker opens up a multitude of possibilities.

\section{Limitations}

First, our method only focuses on maintaining the ID information of a single generated person in the image, and cannot control multiple IDs of generated persons in one image simultaneously.
Second, our method excels at generating half-length portraits, but is relatively not good at generating full-length portraits.
Third, the age transformation ability of our method is not as precise as some GAN-based methods~\cite{alaluf2021matter}. 
If the users need more precise control, modifications to the captions of the training dataset may be required.
Finally, our method is based on the SDXL and the dataset we constructed, so it will also inherit their biases.

\section{Broader Impact}
In this paper, we introduce a novel method capable of generating high-quality human images while maintaining a high degree of similarity to the input identity. 
At the same time, our method can also satisfy high efficiency, decent facial generation diversity, and good controllability.

For the academic community, our method provides a strong baseline for personalized generation. 
Our data creation pipeline enables more diverse datasets with varied poses, actions, and backgrounds, which can be instrumental in developing more robust and generalizable computer vision models.

In the realm of practical applications, our technique has the potential to revolutionize industries such as entertainment, where it can be used to create realistic characters for movies or video games without the need for extensive CGI work.
It can also be beneficial in virtual reality, providing more immersive and personalized experiences by allowing users to see themselves in different scenarios.
It is worth noticing that everyone can rely on our PhotoMaker to quickly customize their own digital portraits.

However, we acknowledge the ethical considerations that arise with the ability to generate human images with high fidelity.
The proliferation of such technology may lead to a surge in the inappropriate use of generated portraits, malicious image tampering, and the spreading of false information.
Therefore, we stress the importance of developing and adhering to ethical guidelines and using this technology responsibly. 
We hope that our contribution will spur further discussion and research into the safe and ethical use of human generation in computer vision.

%% file: arxiv.bbl
\begin{thebibliography}{68}
\providecommand{\natexlab}[1]{#1}
\providecommand{\url}[1]{\texttt{#1}}
\expandafter\ifx\csname urlstyle\endcsname\relax
  \providecommand{\doi}[1]{doi: #1}\else
  \providecommand{\doi}{doi: \begingroup \urlstyle{rm}\Url}\fi

\bibitem[pho()]{photoai}
Photo ai.
\newblock \url{https://photoai.com/}.
\newblock Accessed: 2023-12-08.

\bibitem[db_(2022)]{db_lora}
Low-rank adaptation for fast text-to-image diffusion fine-tuning.
\newblock \url{https://github.com/cloneofsimo/lora}, 2022.

\bibitem[pro(2023)]{prompt_weighting}
Prompt weighting.
\newblock \url{https://huggingface.co/docs/diffusers/using-diffusers/weighted_prompts}, 2023.

\bibitem[Alaluf et~al.(2021)Alaluf, Patashnik, and Cohen-Or]{alaluf2021matter}
Yuval Alaluf, Or Patashnik, and Daniel Cohen-Or.
\newblock Only a matter of style: Age transformation using a style-based regression model.
\newblock \emph{TOG}, 2021.

\bibitem[Arar et~al.(2023)Arar, Gal, Atzmon, Chechik, Cohen-Or, Shamir, and Bermano]{arar2023domain}
Moab Arar, Rinon Gal, Yuval Atzmon, Gal Chechik, Daniel Cohen-Or, Ariel Shamir, and Amit~H Bermano.
\newblock Domain-agnostic tuning-encoder for fast personalization of text-to-image models.
\newblock \emph{TOG}, 2023.

\bibitem[Avrahami et~al.(2023)Avrahami, Aberman, Fried, Cohen-Or, and Lischinski]{avrahami2023break}
Omri Avrahami, Kfir Aberman, Ohad Fried, Daniel Cohen-Or, and Dani Lischinski.
\newblock Break-a-scene: Extracting multiple concepts from a single image.
\newblock In \emph{SIGGRAPH Asia}, 2023.

\bibitem[Cao et~al.(2018)Cao, Shen, Xie, Parkhi, and Zisserman]{cao2018vggface2}
Qiong Cao, Li Shen, Weidi Xie, Omkar~M Parkhi, and Andrew Zisserman.
\newblock Vggface2: A dataset for recognising faces across pose and age.
\newblock In \emph{FG}, 2018.

\bibitem[Caron et~al.(2021)Caron, Touvron, Misra, J{\'e}gou, Mairal, Bojanowski, and Joulin]{caron2021emerging}
Mathilde Caron, Hugo Touvron, Ishan Misra, Herv{\'e} J{\'e}gou, Julien Mairal, Piotr Bojanowski, and Armand Joulin.
\newblock Emerging properties in self-supervised vision transformers.
\newblock In \emph{ICCV}, 2021.

\bibitem[Changpinyo et~al.(2021)Changpinyo, Sharma, Ding, and Soricut]{changpinyo2021conceptual}
Soravit Changpinyo, Piyush Sharma, Nan Ding, and Radu Soricut.
\newblock Conceptual 12m: Pushing web-scale image-text pre-training to recognize long-tail visual concepts.
\newblock In \emph{CVPR}, 2021.

\bibitem[Chen et~al.(2023{\natexlab{a}})Chen, Yu, Ge, Yao, Xie, Wu, Wang, Kwok, Luo, Lu, et~al.]{chen2023pixart}
Junsong Chen, Jincheng Yu, Chongjian Ge, Lewei Yao, Enze Xie, Yue Wu, Zhongdao Wang, James Kwok, Ping Luo, Huchuan Lu, et~al.
\newblock Pixart-alpha: Fast training of diffusion transformer for photorealistic text-to-image synthesis.
\newblock \emph{arXiv preprint arXiv:2310.00426}, 2023{\natexlab{a}}.

\bibitem[Chen et~al.(2023{\natexlab{b}})Chen, Zhao, Liu, Ding, Song, Wang, Wang, Yang, Liu, Du, et~al.]{chen2023photoverse}
Li Chen, Mengyi Zhao, Yiheng Liu, Mingxu Ding, Yangyang Song, Shizun Wang, Xu Wang, Hao Yang, Jing Liu, Kang Du, et~al.
\newblock Photoverse: Tuning-free image customization with text-to-image diffusion models.
\newblock \emph{arXiv preprint arXiv:2309.05793}, 2023{\natexlab{b}}.

\bibitem[Chen et~al.(2023{\natexlab{c}})Chen, Hu, Li, Rui, Jia, Chang, and Cohen]{chen2023subject}
Wenhu Chen, Hexiang Hu, Yandong Li, Nataniel Rui, Xuhui Jia, Ming-Wei Chang, and William~W Cohen.
\newblock Subject-driven text-to-image generation via apprenticeship learning.
\newblock \emph{arXiv preprint arXiv:2304.00186}, 2023{\natexlab{c}}.

\bibitem[Chen et~al.(2023{\natexlab{d}})Chen, Huang, Liu, Shen, Zhao, and Zhao]{chen2023anydoor}
Xi Chen, Lianghua Huang, Yu Liu, Yujun Shen, Deli Zhao, and Hengshuang Zhao.
\newblock Anydoor: Zero-shot object-level image customization.
\newblock \emph{arXiv preprint arXiv:2307.09481}, 2023{\natexlab{d}}.

\bibitem[Cheng et~al.(2022)Cheng, Misra, Schwing, Kirillov, and Girdhar]{cheng2021mask2former}
Bowen Cheng, Ishan Misra, Alexander~G. Schwing, Alexander Kirillov, and Rohit Girdhar.
\newblock Masked-attention mask transformer for universal image segmentation.
\newblock In \emph{CVPR}, 2022.

\bibitem[Deng et~al.(2019)Deng, Guo, Xue, and Zafeiriou]{deng2019arcface}
Jiankang Deng, Jia Guo, Niannan Xue, and Stefanos Zafeiriou.
\newblock Arcface: Additive angular margin loss for deep face recognition.
\newblock In \emph{CVPR}, 2019.

\bibitem[Deng et~al.(2020)Deng, Guo, Ververas, Kotsia, and Zafeiriou]{deng2020retinaface}
Jiankang Deng, Jia Guo, Evangelos Ververas, Irene Kotsia, and Stefanos Zafeiriou.
\newblock Retinaface: Single-shot multi-level face localisation in the wild.
\newblock In \emph{CVPR}, 2020.

\bibitem[Gal et~al.(2023{\natexlab{a}})Gal, Alaluf, Atzmon, Patashnik, Bermano, Chechik, and Cohen-Or]{gal2022image}
Rinon Gal, Yuval Alaluf, Yuval Atzmon, Or Patashnik, Amit~H Bermano, Gal Chechik, and Daniel Cohen-Or.
\newblock An image is worth one word: Personalizing text-to-image generation using textual inversion.
\newblock In \emph{ICLR}, 2023{\natexlab{a}}.

\bibitem[Gal et~al.(2023{\natexlab{b}})Gal, Arar, Atzmon, Bermano, Chechik, and Cohen-Or]{gal2023designing}
Rinon Gal, Moab Arar, Yuval Atzmon, Amit~H Bermano, Gal Chechik, and Daniel Cohen-Or.
\newblock Designing an encoder for fast personalization of text-to-image models.
\newblock \emph{arXiv preprint arXiv:2302.12228}, 2023{\natexlab{b}}.

\bibitem[Goodfellow et~al.(2020)Goodfellow, Pouget-Abadie, Mirza, Xu, Warde-Farley, Ozair, Courville, and Bengio]{goodfellow2020generative}
Ian Goodfellow, Jean Pouget-Abadie, Mehdi Mirza, Bing Xu, David Warde-Farley, Sherjil Ozair, Aaron Courville, and Yoshua Bengio.
\newblock Generative adversarial networks.
\newblock \emph{ACM Communications}, 2020.

\bibitem[Han et~al.(2023)Han, Li, Zhang, Milanfar, Metaxas, and Yang]{han2023svdiff}
Ligong Han, Yinxiao Li, Han Zhang, Peyman Milanfar, Dimitris Metaxas, and Feng Yang.
\newblock Svdiff: Compact parameter space for diffusion fine-tuning.
\newblock In \emph{ICCV}, 2023.

\bibitem[Hertz et~al.(2023)Hertz, Mokady, Tenenbaum, Aberman, Pritch, and Cohen-Or]{hertz2022prompt}
Amir Hertz, Ron Mokady, Jay Tenenbaum, Kfir Aberman, Yael Pritch, and Daniel Cohen-Or.
\newblock Prompt-to-prompt image editing with cross attention control.
\newblock In \emph{ICLR}, 2023.

\bibitem[Heusel et~al.(2017)Heusel, Ramsauer, Unterthiner, Nessler, and Hochreiter]{heusel2017gans}
Martin Heusel, Hubert Ramsauer, Thomas Unterthiner, Bernhard Nessler, and Sepp Hochreiter.
\newblock Gans trained by a two time-scale update rule converge to a local nash equilibrium.
\newblock In \emph{NeurIPS}, 2017.

\bibitem[Ho et~al.(2020)Ho, Jain, and Abbeel]{ho2020denoising}
Jonathan Ho, Ajay Jain, and Pieter Abbeel.
\newblock Denoising diffusion probabilistic models.
\newblock In \emph{NeurIPS}, 2020.

\bibitem[Honnibal et~al.(2020)Honnibal, Montani, Van~Landeghem, and Boyd]{montani20212}
Matthew Honnibal, Ines Montani, Sofie Van~Landeghem, and Adriane Boyd.
\newblock spacy: Industrial-strength natural language processing in python, 2020.

\bibitem[Hu et~al.(2022)Hu, Shen, Wallis, Allen-Zhu, Li, Wang, Wang, and Chen]{hu2021lora}
Edward~J Hu, Yelong Shen, Phillip Wallis, Zeyuan Allen-Zhu, Yuanzhi Li, Shean Wang, Lu Wang, and Weizhu Chen.
\newblock Lora: Low-rank adaptation of large language models.
\newblock In \emph{ICLR}, 2022.

\bibitem[Ilharco et~al.(2021)Ilharco, Wortsman, Wightman, Gordon, Carlini, Taori, Dave, Shankar, Namkoong, Miller, Hajishirzi, Farhadi, and Schmidt]{ilharco_gabriel_2021_5143773}
Gabriel Ilharco, Mitchell Wortsman, Ross Wightman, Cade Gordon, Nicholas Carlini, Rohan Taori, Achal Dave, Vaishaal Shankar, Hongseok Namkoong, John Miller, Hannaneh Hajishirzi, Ali Farhadi, and Ludwig Schmidt.
\newblock Openclip, 2021.

\bibitem[Jia et~al.(2023)Jia, Zhao, Chan, Li, Zhang, Gong, Hou, Wang, and Su]{jia2023taming}
Xuhui Jia, Yang Zhao, Kelvin~CK Chan, Yandong Li, Han Zhang, Boqing Gong, Tingbo Hou, Huisheng Wang, and Yu-Chuan Su.
\newblock Taming encoder for zero fine-tuning image customization with text-to-image diffusion models.
\newblock \emph{arXiv preprint arXiv:2304.02642}, 2023.

\bibitem[Ju et~al.(2023)Ju, Zeng, Zhao, Wang, Zhang, and Xu]{ju2023humansd}
Xuan Ju, Ailing Zeng, Chenchen Zhao, Jianan Wang, Lei Zhang, and Qiang Xu.
\newblock Human{SD}: A native skeleton-guided diffusion model for human image generation.
\newblock In \emph{ICCV}, 2023.

\bibitem[Karras et~al.(2019)Karras, Laine, and Aila]{karras2019style}
Tero Karras, Samuli Laine, and Timo Aila.
\newblock A style-based generator architecture for generative adversarial networks.
\newblock In \emph{CVPR}, 2019.

\bibitem[Kawar et~al.(2023)Kawar, Zada, Lang, Tov, Chang, Dekel, Mosseri, and Irani]{kawar2022imagic}
Bahjat Kawar, Shiran Zada, Oran Lang, Omer Tov, Huiwen Chang, Tali Dekel, Inbar Mosseri, and Michal Irani.
\newblock Imagic: Text-based real image editing with diffusion models.
\newblock In \emph{CVPR}, 2023.

\bibitem[Kingma and Ba(2015)]{kingma2014adam}
Diederik~P Kingma and Jimmy Ba.
\newblock Adam: A method for stochastic optimization.
\newblock In \emph{ICLR}, 2015.

\bibitem[Kumari et~al.(2023)Kumari, Zhang, Zhang, Shechtman, and Zhu]{kumari2022multi}
Nupur Kumari, Bingliang Zhang, Richard Zhang, Eli Shechtman, and Jun-Yan Zhu.
\newblock Multi-concept customization of text-to-image diffusion.
\newblock In \emph{CVPR}, 2023.

\bibitem[Li et~al.(2023{\natexlab{a}})Li, Li, Savarese, and Hoi]{li2023blip}
Junnan Li, Dongxu Li, Silvio Savarese, and Steven Hoi.
\newblock Blip-2: Bootstrapping language-image pre-training with frozen image encoders and large language models.
\newblock In \emph{ICML}, 2023{\natexlab{a}}.

\bibitem[Li et~al.(2023{\natexlab{b}})Li, Liu, Wu, Mu, Yang, Gao, Li, and Lee]{li2023gligen}
Yuheng Li, Haotian Liu, Qingyang Wu, Fangzhou Mu, Jianwei Yang, Jianfeng Gao, Chunyuan Li, and Yong~Jae Lee.
\newblock Gligen: Open-set grounded text-to-image generation.
\newblock In \emph{CVPR}, 2023{\natexlab{b}}.

\bibitem[Liu et~al.(2023)Liu, Ren, Siarohin, Skorokhodov, Li, Lin, Liu, Liu, and Tulyakov]{liu2023hyperhuman}
Xian Liu, Jian Ren, Aliaksandr Siarohin, Ivan Skorokhodov, Yanyu Li, Dahua Lin, Xihui Liu, Ziwei Liu, and Sergey Tulyakov.
\newblock Hyperhuman: Hyper-realistic human generation with latent structural diffusion.
\newblock \emph{arXiv preprint arXiv:2310.08579}, 2023.

\bibitem[Liu et~al.(2015)Liu, Luo, Wang, and Tang]{liu2015faceattributes}
Ziwei Liu, Ping Luo, Xiaogang Wang, and Xiaoou Tang.
\newblock Deep learning face attributes in the wild.
\newblock In \emph{ICCV}, 2015.

\bibitem[Ma et~al.(2023{\natexlab{a}})Ma, Liang, Chen, and Lu]{ma2023subject}
Jian Ma, Junhao Liang, Chen Chen, and Haonan Lu.
\newblock Subject-diffusion: Open domain personalized text-to-image generation without test-time fine-tuning.
\newblock \emph{arXiv preprint arXiv:2307.11410}, 2023{\natexlab{a}}.

\bibitem[Ma et~al.(2023{\natexlab{b}})Ma, Yang, Wang, Fu, and Liu]{ma2023unified}
Yiyang Ma, Huan Yang, Wenjing Wang, Jianlong Fu, and Jiaying Liu.
\newblock Unified multi-modal latent diffusion for joint subject and text conditional image generation.
\newblock \emph{arXiv preprint arXiv:2303.09319}, 2023{\natexlab{b}}.

\bibitem[Melnik et~al.(2022)Melnik, Miasayedzenkau, Makarovets, Pirshtuk, Akbulut, Holzmann, Renusch, Reichert, and Ritter]{melnik2022face}
Andrew Melnik, Maksim Miasayedzenkau, Dzianis Makarovets, Dzianis Pirshtuk, Eren Akbulut, Dennis Holzmann, Tarek Renusch, Gustav Reichert, and Helge Ritter.
\newblock Face generation and editing with stylegan: A survey.
\newblock \emph{arXiv preprint arXiv:2212.09102}, 2022.

\bibitem[Mou et~al.(2023)Mou, Wang, Xie, Zhang, Qi, Shan, and Qie]{mou2023t2i}
Chong Mou, Xintao Wang, Liangbin Xie, Jian Zhang, Zhongang Qi, Ying Shan, and Xiaohu Qie.
\newblock T2i-adapter: Learning adapters to dig out more controllable ability for text-to-image diffusion models.
\newblock \emph{arXiv preprint arXiv:2302.08453}, 2023.

\bibitem[Nitzan et~al.(2022)Nitzan, Aberman, He, Liba, Yarom, Gandelsman, Mosseri, Pritch, and Cohen-Or]{nitzan2022mystyle}
Yotam Nitzan, Kfir Aberman, Qiurui He, Orly Liba, Michal Yarom, Yossi Gandelsman, Inbar Mosseri, Yael Pritch, and Daniel Cohen-Or.
\newblock Mystyle: A personalized generative prior.
\newblock \emph{TOG}, 2022.

\bibitem[Parmar et~al.(2022)Parmar, Zhang, and Zhu]{parmar2021cleanfid}
Gaurav Parmar, Richard Zhang, and Jun-Yan Zhu.
\newblock On aliased resizing and surprising subtleties in gan evaluation.
\newblock In \emph{CVPR}, 2022.

\bibitem[Peebles and Xie(2023)]{peebles2023scalable}
William Peebles and Saining Xie.
\newblock Scalable diffusion models with transformers.
\newblock In \emph{ICCV}, 2023.

\bibitem[Podell et~al.(2023)Podell, English, Lacey, Blattmann, Dockhorn, M{\"u}ller, Penna, and Rombach]{podell2023sdxl}
Dustin Podell, Zion English, Kyle Lacey, Andreas Blattmann, Tim Dockhorn, Jonas M{\"u}ller, Joe Penna, and Robin Rombach.
\newblock Sdxl: Improving latent diffusion models for high-resolution image synthesis.
\newblock \emph{arXiv preprint arXiv:2307.01952}, 2023.

\bibitem[Radford et~al.(2021)Radford, Kim, Hallacy, Ramesh, Goh, Agarwal, Sastry, Askell, Mishkin, Clark, et~al.]{radford2021learning}
Alec Radford, Jong~Wook Kim, Chris Hallacy, Aditya Ramesh, Gabriel Goh, Sandhini Agarwal, Girish Sastry, Amanda Askell, Pamela Mishkin, Jack Clark, et~al.
\newblock Learning transferable visual models from natural language supervision.
\newblock In \emph{ICML}, 2021.

\bibitem[Raffel et~al.(2020)Raffel, Shazeer, Roberts, Lee, Narang, Matena, Zhou, Li, and Liu]{raffel2020exploring}
Colin Raffel, Noam Shazeer, Adam Roberts, Katherine Lee, Sharan Narang, Michael Matena, Yanqi Zhou, Wei Li, and Peter~J Liu.
\newblock Exploring the limits of transfer learning with a unified text-to-text transformer.
\newblock \emph{JMLR}, 2020.

\bibitem[Ramesh et~al.(2022)Ramesh, Dhariwal, Nichol, Chu, and Chen]{ramesh2022hierarchical}
Aditya Ramesh, Prafulla Dhariwal, Alex Nichol, Casey Chu, and Mark Chen.
\newblock Hierarchical text-conditional image generation with clip latents.
\newblock \emph{arXiv preprint arXiv:2204.06125}, 2022.

\bibitem[Reimers and Gurevych(2019)]{reimers2019sentence}
Nils Reimers and Iryna Gurevych.
\newblock Sentence-bert: Sentence embeddings using siamese bert-networks.
\newblock In \emph{EMNLP}, 2019.

\bibitem[Rombach et~al.(2022)Rombach, Blattmann, Lorenz, Esser, and Ommer]{rombach2022high}
Robin Rombach, Andreas Blattmann, Dominik Lorenz, Patrick Esser, and Bj{\"o}rn Ommer.
\newblock High-resolution image synthesis with latent diffusion models.
\newblock In \emph{CVPR}, 2022.

\bibitem[Ruiz et~al.(2023{\natexlab{a}})Ruiz, Li, Jampani, Pritch, Rubinstein, and Aberman]{ruiz2022dreambooth}
Nataniel Ruiz, Yuanzhen Li, Varun Jampani, Yael Pritch, Michael Rubinstein, and Kfir Aberman.
\newblock Dreambooth: Fine tuning text-to-image diffusion models for subject-driven generation.
\newblock In \emph{CVPR}, 2023{\natexlab{a}}.

\bibitem[Ruiz et~al.(2023{\natexlab{b}})Ruiz, Li, Jampani, Wei, Hou, Pritch, Wadhwa, Rubinstein, and Aberman]{ruiz2023hyperdreambooth}
Nataniel Ruiz, Yuanzhen Li, Varun Jampani, Wei Wei, Tingbo Hou, Yael Pritch, Neal Wadhwa, Michael Rubinstein, and Kfir Aberman.
\newblock Hyperdreambooth: Hypernetworks for fast personalization of text-to-image models.
\newblock \emph{arXiv preprint arXiv:2307.06949}, 2023{\natexlab{b}}.

\bibitem[Saharia et~al.(2022)Saharia, Chan, Saxena, Li, Whang, Denton, Ghasemipour, Ayan, Mahdavi, Lopes, et~al.]{saharia2022photorealistic}
Chitwan Saharia, William Chan, Saurabh Saxena, Lala Li, Jay Whang, Emily Denton, Seyed Kamyar~Seyed Ghasemipour, Burcu~Karagol Ayan, S~Sara Mahdavi, Rapha~Gontijo Lopes, et~al.
\newblock Photorealistic text-to-image diffusion models with deep language understanding.
\newblock In \emph{NeurIPS}, 2022.

\bibitem[Schroff et~al.(2015)Schroff, Kalenichenko, and Philbin]{schroff2015facenet}
Florian Schroff, Dmitry Kalenichenko, and James Philbin.
\newblock Facenet: A unified embedding for face recognition and clustering.
\newblock In \emph{CVPR}, 2015.

\bibitem[Schuhmann et~al.(2021)Schuhmann, Vencu, Beaumont, Kaczmarczyk, Mullis, Katta, Coombes, Jitsev, and Komatsuzaki]{schuhmann2021laion}
Christoph Schuhmann, Richard Vencu, Romain Beaumont, Robert Kaczmarczyk, Clayton Mullis, Aarush Katta, Theo Coombes, Jenia Jitsev, and Aran Komatsuzaki.
\newblock Laion-400m: Open dataset of clip-filtered 400 million image-text pairs.
\newblock \emph{arXiv preprint arXiv:2111.02114}, 2021.

\bibitem[Schuhmann et~al.(2022)Schuhmann, Beaumont, Vencu, Gordon, Wightman, Cherti, Coombes, Katta, Mullis, Wortsman, et~al.]{schuhmann2022laion}
Christoph Schuhmann, Romain Beaumont, Richard Vencu, Cade Gordon, Ross Wightman, Mehdi Cherti, Theo Coombes, Aarush Katta, Clayton Mullis, Mitchell Wortsman, et~al.
\newblock Laion-5b: An open large-scale dataset for training next generation image-text models.
\newblock \emph{arXiv preprint arXiv:2210.08402}, 2022.

\bibitem[Shi et~al.(2023)Shi, Xiong, Lin, and Jung]{shi2023instantbooth}
Jing Shi, Wei Xiong, Zhe Lin, and Hyun~Joon Jung.
\newblock Instantbooth: Personalized text-to-image generation without test-time finetuning.
\newblock \emph{arXiv preprint arXiv:2304.03411}, 2023.

\bibitem[Sohn et~al.(2023)Sohn, Ruiz, Lee, Chin, Blok, Chang, Barber, Jiang, Entis, Li, et~al.]{sohn2023styledrop}
Kihyuk Sohn, Nataniel Ruiz, Kimin Lee, Daniel~Castro Chin, Irina Blok, Huiwen Chang, Jarred Barber, Lu Jiang, Glenn Entis, Yuanzhen Li, et~al.
\newblock Styledrop: Text-to-image generation in any style.
\newblock In \emph{NeurIPS}, 2023.

\bibitem[Song et~al.(2021)Song, Meng, and Ermon]{song2020denoising}
Jiaming Song, Chenlin Meng, and Stefano Ermon.
\newblock Denoising diffusion implicit models.
\newblock In \emph{ICLR}, 2021.

\bibitem[Wang et~al.(2018)Wang, Zheng, Liang, Chen, Lin, and Yang]{wang2018toward}
Bochao Wang, Huabin Zheng, Xiaodan Liang, Yimin Chen, Liang Lin, and Meng Yang.
\newblock Toward characteristic-preserving image-based virtual try-on network.
\newblock In \emph{ECCV}, 2018.

\bibitem[Wang et~al.(2020)Wang, Wu, Song, Yang, Wu, Qian, He, Qiao, and Loy]{kaisiyuan2020mead}
Kaisiyuan Wang, Qianyi Wu, Linsen Song, Zhuoqian Yang, Wayne Wu, Chen Qian, Ran He, Yu Qiao, and Chen~Change Loy.
\newblock Mead: A large-scale audio-visual dataset for emotional talking-face generation.
\newblock In \emph{ECCV}, 2020.

\bibitem[Wei et~al.(2023)Wei, Zhang, Ji, Bai, Zhang, and Zuo]{wei2023elite}
Yuxiang Wei, Yabo Zhang, Zhilong Ji, Jinfeng Bai, Lei Zhang, and Wangmeng Zuo.
\newblock Elite: Encoding visual concepts into textual embeddings for customized text-to-image generation.
\newblock In \emph{ICCV}, 2023.

\bibitem[Xiao et~al.(2023)Xiao, Yin, Freeman, Durand, and Han]{xiao2023fastcomposer}
Guangxuan Xiao, Tianwei Yin, William~T Freeman, Fr{\'e}do Durand, and Song Han.
\newblock Fastcomposer: Tuning-free multi-subject image generation with localized attention.
\newblock \emph{arXiv preprint arXiv:2305.10431}, 2023.

\bibitem[Ye et~al.(2023)Ye, Zhang, Liu, Han, and Yang]{ye2023ip}
Hu Ye, Jun Zhang, Sibo Liu, Xiao Han, and Wei Yang.
\newblock Ip-adapter: Text compatible image prompt adapter for text-to-image diffusion models.
\newblock \emph{arXiv preprint arXiv:2308.06721}, 2023.

\bibitem[Yuan et~al.(2023)Yuan, Cun, Zhang, Li, Qi, Wang, Shan, and Zheng]{yuan2023inserting}
Ge Yuan, Xiaodong Cun, Yong Zhang, Maomao Li, Chenyang Qi, Xintao Wang, Ying Shan, and Huicheng Zheng.
\newblock Inserting anybody in diffusion models via celeb basis.
\newblock In \emph{NeurIPS}, 2023.

\bibitem[Zhang et~al.(2023{\natexlab{a}})Zhang, Rao, and Agrawala]{zhang2023adding}
Lvmin Zhang, Anyi Rao, and Maneesh Agrawala.
\newblock Adding conditional control to text-to-image diffusion models.
\newblock In \emph{ICCV}, 2023{\natexlab{a}}.

\bibitem[Zhang et~al.(2018)Zhang, Isola, Efros, Shechtman, and Wang]{zhang2018perceptual}
Richard Zhang, Phillip Isola, Alexei~A Efros, Eli Shechtman, and Oliver Wang.
\newblock The unreasonable effectiveness of deep features as a perceptual metric.
\newblock In \emph{CVPR}, 2018.

\bibitem[Zhang et~al.(2023{\natexlab{b}})Zhang, Cun, Wang, Zhang, Shen, Guo, Shan, and Wang]{zhang2023sadtalker}
Wenxuan Zhang, Xiaodong Cun, Xuan Wang, Yong Zhang, Xi Shen, Yu Guo, Ying Shan, and Fei Wang.
\newblock Sadtalker: Learning realistic 3d motion coefficients for stylized audio-driven single image talking face animation.
\newblock In \emph{CVPR}, 2023{\natexlab{b}}.

\bibitem[Zheng et~al.(2022)Zheng, Yang, Zhang, Bao, Chen, Huang, Yuan, Chen, Zeng, and Wen]{zheng2022general}
Yinglin Zheng, Hao Yang, Ting Zhang, Jianmin Bao, Dongdong Chen, Yangyu Huang, Lu Yuan, Dong Chen, Ming Zeng, and Fang Wen.
\newblock General facial representation learning in a visual-linguistic manner.
\newblock In \emph{CVPR}, 2022.

\end{thebibliography}
